\documentclass[acmsmall]{acmart}
\AtBeginDocument{%
  }

\setcopyright{acmlicensed}
\copyrightyear{2018}
\acmYear{2018}
\acmDOI{XXXXXXX.XXXXXXX}

\usepackage{multirow}
\usepackage{graphicx}
\usepackage[table,xcdraw]{xcolor}
\usepackage{colortbl}
\usepackage{url}

\definecolor{myred}{rgb}{1,0,0}
\definecolor{mygreen}{rgb}{0,0.69,0.314}
\definecolor{myhighlight}{rgb}{1,0.937,0.918}
\definecolor{mygray}{rgb}{0.929,0.929,0.929}

\newcommand{\redtext}[1]{\textcolor{myred}{\textit{#1}}}

\usepackage{tcolorbox}
\tcbuselibrary{most}     
\tcbuselibrary{skins}    
\usepackage{setspace}

\usepackage{graphicx}
\usepackage{wrapfig}
\usepackage{bbm}
\usepackage{url}
\usepackage{algorithm}
\usepackage{textcomp}
\usepackage{amsmath}
\usepackage[caption=false,farskip=0pt,labelfont={bf}]{subfig}
\usepackage[T1]{fontenc}
\usepackage{booktabs} 
\usepackage{xcolor}
\usepackage{multirow}
\usepackage{color}
\usepackage{wrapfig}
\begin{document}

\title{LogReasoner: 
Empowering LLMs with Expert-like Coarse-to-Fine Reasoning for Automated Log Analysis}

\author{Lipeng Ma}
\email{lpma21@m.fudan.edu.cn}
\orcid{0000-0001-5974-5988}
\affiliation{%
  \institution{Shanghai Key Laboratory of Data Science, School of Computer Science, Fudan University}
  \city{Shanghai}
  \country{China}
}

\author{Yixuan Li}
\orcid{0000-0002-9229-7555}
\affiliation{%
 \institution{Shanghai Key Laboratory of Data Science, School of Computer Science, Fudan University}
 \city{Shanghai}
 \country{China}
}
\email{yxli24@m.fudan.edu.cn}

\author{Weidong Yang}
\orcid{0000-0002-6473-9272}
\affiliation{%
 \institution{Shanghai Key Laboratory of Data Science, School of Computer Science, Fudan University}
 \city{Shanghai}
 \country{China}
}
\email{wdyang@fudan.edu.cn}

\author{Mingjie Zhou}
\orcid{0000-0002-3289-0533}
\affiliation{%
 \institution{Shanghai Key Laboratory of Data Science, School of Computer Science, Fudan University}
 \city{Shanghai}
 \country{China}
 \postcode{200433}
}
\email{mjzhou19@fudan.edu.cn}

\author{Xinyi Liu}
\affiliation{%
 \institution{Shanghai Key Laboratory of Data Science, School of Computer Science, Fudan University}
 \city{Shanghai}
 \country{China}
 \postcode{200433}
}
\email{liuxiny24@m.fudan.edu.cn}

\author{Ben Fei}
\orcid{0000-0002-3219-9996}
\affiliation{%
 \institution{Shanghai Key Laboratory of Data Science, School of Computer Science, Fudan University}
 \city{Shanghai}
 \country{China}
 \postcode{200433}
}
\email{bfei21@m.fudan.edu.cn}

\author{Shuhao Li}
\orcid{0009-0008-5175-7667}
\affiliation{%
 \institution{Shanghai Key Laboratory of Data Science, School of Computer Science, Fudan University}
 \city{Shanghai}
 \country{China}
 \postcode{200433}
}
\email{shli23@m.fudan.edu.cn}

\author{Xiaoyan Sun}
\affiliation{%
 \institution{School of Artificial Intelligence and Computer Science, Jiangnan University}
 \city{Wuxi}
 \country{China}
 \postcode{200433}
}
\email{xysun78@jiangnan.edu.cn}

\author{Sihang Jiang}
\orcid{0000-0002-0736-6457}
\affiliation{%
 \institution{Shanghai Key Laboratory of Data Science, School of Computer Science, Fudan University}
 \city{Shanghai}
 \country{China}
 \postcode{200433}
}
\email{jiangsihang@fudan.edu.cn}

\author{Yanghua Xiao}
\orcid{0000-0001-8403-9591}
\affiliation{%
 \institution{Shanghai Key Laboratory of Data Science, School of Computer Science, Fudan University}
 \city{Shanghai}
 \country{China}
 \postcode{200433}
}
\email{shawyh@fudan.edu.cn}

\renewcommand{\shortauthors}{L. Ma et al.}

\begin{abstract}
Log analysis is crucial for monitoring system health and diagnosing failures in complex systems. Recent advances in large language models (LLMs) offer new opportunities for automated log analysis, leveraging their reasoning capabilities to perform tasks such as anomaly detection and failure prediction. 
However, general-purpose LLMs struggle to formulate structured reasoning workflows that align with expert cognition and deliver precise details of reasoning steps.
To address these challenges, we propose LogReasoner, a coarse-to-fine reasoning enhancement framework designed to enable LLMs to reason log analysis tasks like experts. 
LogReasoner consists of two stages: (1) coarse-grained enhancement of expert thinking, where high-level expert thoughts are constructed from collected troubleshooting flowcharts and existing tasks to enable LLMs to formulate structured reasoning workflows and (2) fine-grained enhancement of specific steps, where we first fine-tune the LLM with task-specific stepwise solutions to enhance the LLM for instantiated reasoning, then employ the preference learning to calibrate the LLM's reasoning details from its mistakes, further strengthen the LLM's analytical granularity and correctness.
We evaluate LogReasoner on four distinct log analysis tasks using open-source LLMs such as Qwen-2.5 and Llama-3. Experimental results show that LogReasoner significantly outperforms existing LLMs, achieving state-of-the-art performance and demonstrating its effectiveness in enhancing the reasoning capabilities of LLMs for log analysis.
Our source code and detailed experimental data are available at \url{https://github.com/LeaperOvO/LogReasoner}.
\end{abstract}



\maketitle

\section{Introduction}

\begin{figure}[]
  \centering
  \includegraphics[width=0.88\columnwidth]{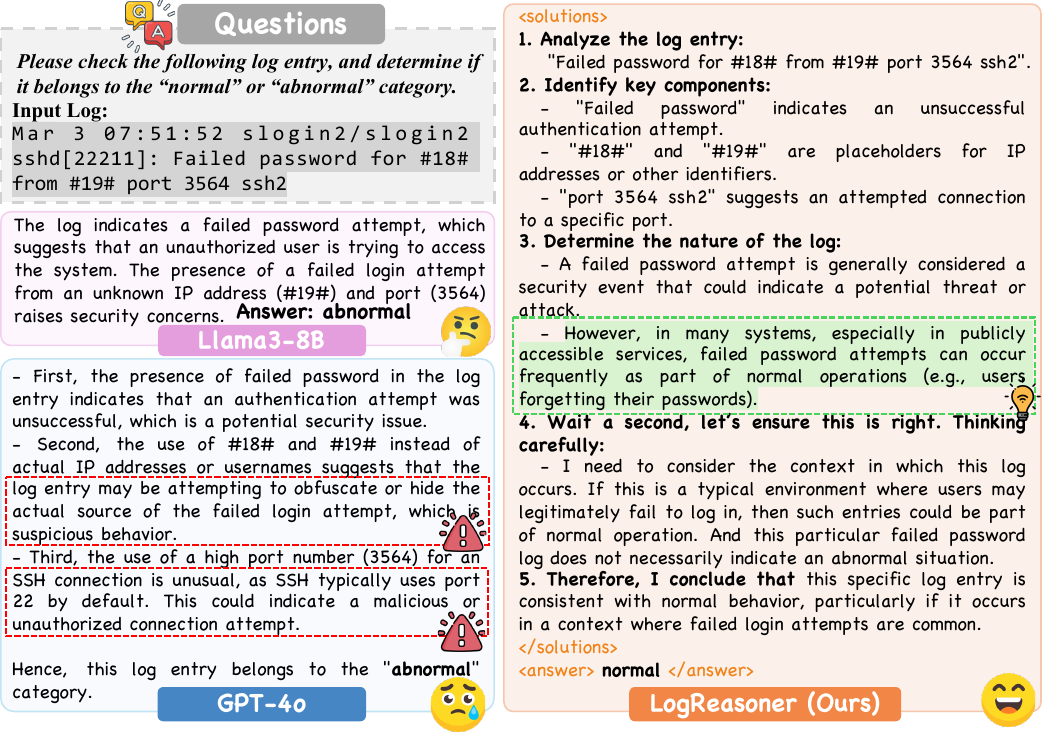}
  \caption{Examples of different LLMs reasoning on anomaly detection, where Llama3 and GPT-4o lack the expert thinking to formulate a reasoning workflow and provide precise evidence within the reasoning steps. In contrast, LogReasoner explicitly decomposes the task by first outlining a high-level reasoning framework, then filling it with precise details, and ultimately generating a reliable answer.}
  \label{fig:reason_examples}
\end{figure}

Logs are indispensable for recording the runtime status of complex systems, providing crucial information for monitoring system health and diagnosing failures  \cite{he2021survey,zhang2024failure,cui2024logeval}. Traditionally, log analysis has relied heavily on experienced specialists manually inspecting massive volumes of log data,  a process that is both labor-intensive and inefficient \cite{zhang2021onion}.
To address these inefficiencies, automated log analysis has emerged as a promising solution, leveraging machine learning or deep learning models to identify critical information from log data \cite{du2017deeplog,zhang2019robust,li2020swisslog,le2021log,liu2022uniparser,almodovar2024logfit,le2022log,yu2024deep,zhang2024metalog}.
Recently, driven by the rapid advances in large language models (LLMs) and their enhanced reasoning abilities \cite{chen2025towards}, LLMs have evolved beyond simple answer generation to simulate human-like, step-by-step reasoning processes \cite{wei2022chain,chen2025towards}. This has led to remarkable success in solving complex problems, such as code generation \cite{du2024evaluating,jiang2024self} and mathematical reasoning \cite{huang2025key,xia2025evaluating}.
Inspired by this, LLM-based approaches for automated log analysis have attracted widespread attention \cite{le2023log,jiang2023lilac,xu2024divlog,ma2024knowlog,zhong2024logparser,ma2024luk,liu2024loglm,wang2024logexpert,ma2025adaptivelog,xuopenrca,astekin2024comparative,le2025unleashing}.
Unlike traditional methods, which typically require extensive labeled data and domain-specific training, LLMs can analyze logs via prompting techniques, offering parameter-free solutions and achieving significant progress in tasks such as anomaly detection and failure prediction.

Despite these advancements, general-purpose LLMs still face significant gaps when applied to real-world log analysis tasks that require intricate reasoning and domain-specific expertise.



First, general-purpose LLMs struggle to formulate structured reasoning workflows that align with expert cognition.
Real-world log analysis typically requires multi-step reasoning, where engineers decompose problems into a sequence of logical steps to guide the analysis process.
For instance, experienced engineers begin by sketching a diagnostic workflow based on observed symptoms and then follow it systematically to identify root causes. This structured progression ensures consistency and logical coherence throughout troubleshooting.
Unfortunately, general-purpose LLMs frequently produce fragmented or unfocused reasoning processes. This limitation arises due to their pre-training on open-domain data, such as encyclopedias and programming code, with limited exposure to domain-specific corpora, such as system logs, operational reports, or diagnostic flowcharts.
Consequently, LLMs often fail to summarize systematic analytical workflows that focus on how to reason on log analysis tasks. For example, as shown in Figure \ref{fig:reason_examples}, when analyzing a failed SSH login, Llama3 and GPT-4o generate unrelated steps lacking clear logical progression, failing to establish a coherent diagnostic flow.

Second, LLMs are prone to errors in handling the detailed information embedded within logs.
Unlike natural language, logs typically contain dynamic variables (e.g., timestamps, ports, IP addresses) and diverse symbolic formats \cite{ma2024knowlog}. LLMs frequently make mistakes in processing these unfamiliar details or overlook critical information during reasoning. Log analysis is highly detail-sensitive, and such oversights can lead to significant reasoning gaps in practical applications. 
For example, as shown in Figure \ref{fig:reason_examples}, when determining whether a failed SSH login is abnormal, GPT-4o disproportionately focuses on superficial indicators like anonymized fields or unusual port numbers, treating these isolated details as abnormal without synthesizing critical contextual evidence. This lack of precision compromises the accuracy and reliability of the conclusions, undermining trust in LLM-generated results.

Critically, these two limitations exhibit a strong interdependence, where the structured workflow provides the essential context and logical sequencing necessary to interpret specific log details accurately. Conversely, the accuracy of details affects the progression of the subsequent reasoning steps. 
Consequently, this paper focuses not only on the expert thinking but also on the reasoning details.

To address these challenges, we propose LogReasoner, a reasoning enhancement framework from coarse-grained to fine-grained enables LLMs to reason log analysis tasks like experts. 
Specifically, LogReasoner first enhances the reasoning workflows of LLMs with high-level expert thoughts from the coarse-grained. These high-level thoughts abstract structured logic steps from collected troubleshooting flowcharts and existing tasks by prompting an advanced LLM, forming actionable guidelines for similar problems. 
Then, LogReasoner enhances the details within the reasoning steps from the fine-grained by Imitation Learning and Steps of Calibration.
Imitation Learning collects correct solutions and leverages them for Supervised Fine-Tuning (SFT), enhancing the LLM to follow the structured workflows to instantiate reasoning. And the Steps of Calibration further calibrate the LLM’s reasoning details from its mistakes with Direct Preference Optimization (DPO).

To validate the effectiveness of LogReasoner, we conduct extensive experiments across four distinct log analysis tasks using multiple open-source LLMs, including Qwen-2.5 and Llama-3. The results demonstrate that LogReasoner achieves state-of-the-art performance compared to existing LLMs. For instance, in the anomaly detection task, the enhanced Llama3-8B achieves an average F1-score improvement of 18.93\% over the original Llama3-8B. Additionally, it outperforms ChatGPT and GPT-4o by 17.78\% and 14.65\%, respectively. These results highlight the effectiveness of LogReasoner in significantly improving the reasoning capabilities of LLMs for log analysis tasks.

Our main contributions can be summarized as follows:
\begin{enumerate}
    \item To our best knowledge, we are the first to introduce the reasoning enhancement framework specifically designed for LLM-based log analysis, dubbed LogReasoner, which guides LLMs through a hierarchical process and significantly improves the reasoning ability of LLMs to analyze logs with expert-like coherence and accuracy.

    \item We propose to enhance log reasoning from coarse-grained to fine-grained. First, at the coarse-grained level, we enhance the reasoning workflow by abstracting high-level expert thoughts from diagnostic flowcharts to provide a structured, logical sequence. Second, at the fine-grained level, we enhance the details of each reasoning step through Imitation Learning, followed by Steps of Calibration to correct errors and refine precision.

    \item LogReasoner achieves state-of-the-art results on four different log analysis tasks compared to existing LLMs, notably enabling the enhanced Llama3-8B to surpass powerful counterparts like ChatGPT and GPT-4o by a significant margin. This conclusively validates the practical effectiveness of our approach in bridging the reasoning gap for real-world log analysis.
    
\end{enumerate}

\section{Related Works}
\subsection{Large Language Models for Log Analysis}
The rapid advancements in LLMs have unlocked significant potential for automated log analysis, tackling a range of tasks from foundational log parsing to more complex analytical tasks.
Early research has shown that LLMs are capable of performing foundational log parsing with remarkable accuracy.
For example, LLMParser \cite{ma2024llmparser} demonstrates that the in-context learning abilities of LLMs allow them to surpass traditional log parsers, even in the absence of extensive domain adaptation. Other solutions, such as LogPrompt \cite{liu2024interpretable}, DivLog \cite{xu2024divlog}, LILAC \cite{jiang2023lilac}, SelfLog \cite{pei2024self}, LogBatcher \cite{xiao2024free}, and AdaParser \cite{wu2025log} have further established that LLMs can efficiently process both structured and semi-structured log data, reducing reliance on manual feature engineering.
Beyond parsing, LLMs are increasingly being leveraged for complex analytical tasks within log analysis. Recent frameworks, including UniLog \cite{xu2024unilog}, LogLLM \cite{guan2024logllm}, LogExpert \cite{wang2024logexpert}, AdaptiveLog \cite{ma2025adaptivelog}, ScalaLog \cite{zhang2025scalalog}, LogInsight \cite{sun2025accurate}, and OpenRCA \cite{xuopenrca}, have applied LLMs to complex scenarios such as anomaly detection, failure prediction, and root cause analysis. 
These studies indicate that LLMs excel at extracting and understanding log information \cite{ma2024luk,jiang2024large} and achieve near-perfect results on public benchmarks for log parsing \cite{zhu2023loghub}.

However, despite these advances, LLMs still exhibit notable limitations in tasks that require deep, multi-step reasoning. Their reasoning processes often lack the structure and consistency characteristic of human experts, limiting practical applicability. 
To bridge this gap, this work introduces a coarse-to-fine reasoning enhancement framework designed to bolster the reasoning skills of LLM for log analysis systematically.

\subsection{Reasoning Enhancement in LLMs}
LLMs have demonstrated remarkable reasoning capabilities \cite{achiam2023gpt,guo2025deepseek}, primarily attributed to their adequate corpus and autoregressive nature. 
Prior research has explored enhancing reasoning capabilities by generating intermediate processes, like the Chain-of-Thought (CoT) framework \cite{wei2022chain,chen2025towards}. By encouraging models to think intermediate reasoning steps before arriving at a final answer, these reasoning-based methods have achieved significant improvements in tasks such as mathematical problem-solving \cite{huang2025key,xia2025evaluating} and code generation \cite{du2024evaluating,jiang2024self}.
However, these models remain struggling with reasoning-intensive tasks, particularly when faced with complex problems, often generating logically inconsistent or unsupported conclusions \cite{tyen2024llms}.
More recent advancements have focused on leveraging reinforcement learning (RL) techniques \cite{ouyang2022training,guo2025deepseek} to optimize reasoning in LLMs, enabling LLMs to solve complex problems effectively. RL-based approaches are computationally expensive and require significant engineering effort, limiting their scalability and practicality.  To reduce this complexity, Direct Preference Optimization (DPO) \cite{rafailov2023direct} has been proposed as an alternative, using pairwise preference data for optimization without the significant overhead.

Despite these general advances, existing reasoning enhancement methods have not been adequately tailored to the unique demands of log analysis.  Even recent reasoning-optimized models like DeepSeek-R1~\cite{guo2025deepseek} lack integration with expert troubleshooting knowledge essential for reliable log analysis. Our work fills this gap by introducing LogReasoner, a novel framework that injects expert-inspired reasoning structure and detailed calibration into LLMs for robust and accurate log analysis.

\section{Methods}

\subsection{Overview}

\begin{figure*}[h]
  \centering
  \includegraphics[width=\textwidth]{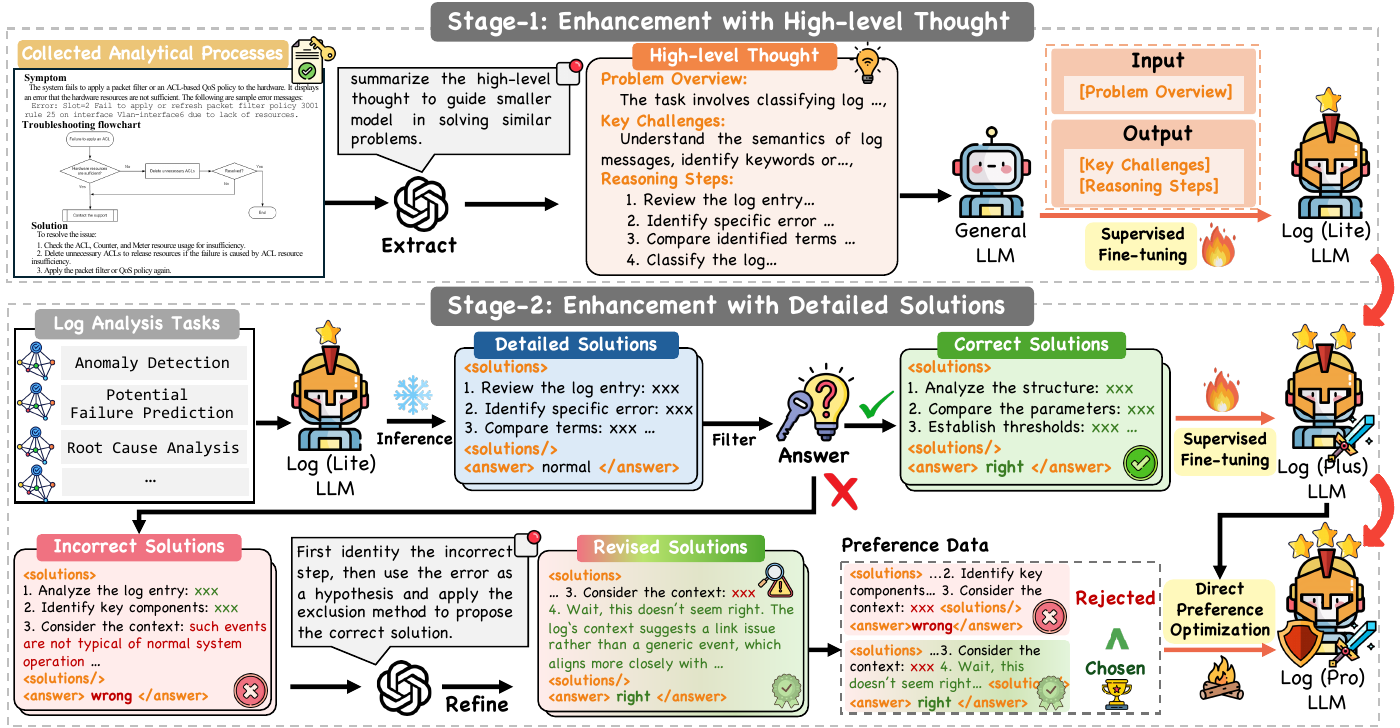}
  \caption{Framework of LogReasoner, which comprises two stages: reasoning workflow enhancement and details enhancement.}
  \label{fig:reason_framework}
\end{figure*}

Fig. \ref{fig:reason_framework} illustrates the overall framework of LogReasoner, which consists of two stages.
Specifically, in the first stage, LogReasoner focuses on the reasoning workflow enhancement. We begin by constructing structured high-level thoughts, which are used to supervise fine-tuning the open-source LLM, enabling LLMs to develop structured workflow planning capabilities from the coarse-grained.
In the second stage,  LogReasoner focuses on the details of reasoning steps from the fine-grained. First, the thought-enhanced LLM from the first stage generates stepwise solutions for specific log analysis tasks, and correct solutions are collected to enhance the LLM for instantiated reasoning with SFT. Second, for erroneous solutions, we identify and refine the flawed reasoning step, and then these corrected solutions are leveraged to teach the LLM to learn from its mistakes with DPO.
In the following sections, we describe the details of LogReasoner.



\subsection{Enhancement with High-Level Thought from Coarse-grained}
Conventional training datasets often drive LLMs to imitate provided results while neglecting the high-level thought. In contrast, experts typically summarize high-level guidelines or workflows before engaging in concrete troubleshooting steps, which serve as a foundation for specific troubleshooting.
Inspired by this, we enhance LLMs with high-level expert thinking to facilitate the formulation of stepwise reasoning workflows.

\subsubsection{High-Level Thought Template Construction}
High-level expert thoughts refer to the abstract problem-solving frameworks and generalized reasoning logic that experts develop before executing concrete steps, which follow three stages to construct.

(1) Raw Analytical Processes Collection: We first gather a diverse set of log analysis cases and their analytical processes from two primary sources: the public maintenance handbook \footnote{\url{https://www.h3c.com/en/Support/Resource_Center/EN/Switches/Catalog/S12500X-AF/S12500X-AF/Technical_Documents/Diagnose___Maintain/Troubleshooting/H3C_S12500-X_S12500X-AF_TG-6W100/}}  and existing log analysis tasks. 
We collect expert-constructed troubleshooting flowcharts of different fault cases and their corresponding explanations from maintenance handbooks released by vendors (e.g., Huawei and H3C) as original analysis processes, where the symptom descriptions as input problems and the explanations of flowcharts as raw analyses.
To ensure data quality, we exclude cases with overly brief symptom descriptions or flowcharts with fewer than three steps.

\begin{figure}[]
  \centering
  \includegraphics[width=0.75\columnwidth]{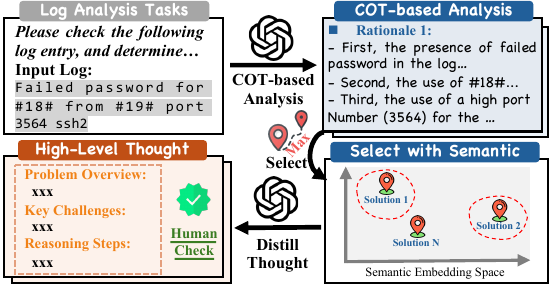}
  \caption{Pipeline of high-level thought construction with existing datasets.}
  \label{fig:reason_thought}
  
\end{figure}

Considering the difficulty of collecting flowcharts for various log analysis tasks in practice, to further increase the diversity of thought, we also incorporate four representative log analysis tasks (anomaly detection, log semantic matching, failure prediction, and root cause analysis) and select samples from publicly available datasets.
As shown in Figure \ref{fig:reason_thought}, then employ GPT-4o to generate CoT-based rationales for these instances, retaining only those chains that lead to correct answers as raw analytical processes. To encourage diverse reasoning, the LLM’s temperature is set to 0.8.

(2) Semantic Filter: To ensure diversity of high-level thoughts, we reduce redundancy in the collected raw data. Specifically, we first take the pre-trained embedding model to encode the semantic representations of all collected analyses.
Then, following practices in DivLog \cite{xu2024divlog}, we employ the Determinantal Point Process (DPP) algorithm \cite{chen2018fast} to select a subset of raw analyses that are maximally diverse in semantic space. Finally, we collect a set $D = \{(x,\hat{s})\}$ of log analysis problems $x$ with raw analytical processes $\hat{s}$.

(3) Template Construction: Given the curated set of problem-analysis pairs $(x,\hat{s})$, we further abstract each analytical process $\hat{s}$ into a high-level thought template $\mathbb{T}_{temp} = (C_{high}, T_{high})$. Specifically, for each input pair, the powerful LLM is prompted to identify the core challenge $C_{high}$ of the input problem and to distill a generalized problem-solving framework consisting of high-level action steps $T_{high}$ without including specific implementation details. 
This abstraction aims to organize reusable reasoning logic and reasoning workflows from expert experience applicable to similar log analysis scenarios. 
Given the tendency of LLMs to produce hallucinated outputs, all generated templates undergo manual review and refinement to ensure quality and rationality. Then we can obtain the high-quality high-level thought dataset $D_{thought}$ as:
\begin{equation}
    D_{thought} = LLM_{Distill}(x,\hat{s}) = \{x,C_{high},T_{high} | x \in D\}, 
\end{equation}
where $T_{high}$ is the formalized guidance trajectory that does not contain specific details.


\subsubsection{Thought-based Supervised Fine-tuning}
Building on our curated high-level thought dataset $D_{thought}$, we employ SFT to equip LLMs with explicit reasoning framework planning capabilities for log analysis. The objective of this phase is to teach the LLM $\pi$ how to think, recognizing the key analytical challenges and, more importantly, formulating the corresponding high-level reasoning framework. Specifically, for a given log analysis input $x$, the model is trained to generate both the core challenge $C_{high}$ and the reasoning framework $T_{high}$. It can be formulated as:
\begin{equation}
\mathcal{L}_{thought} = argmax \sum_{(x,C_{high},T_{high}) \in D_{thought}} log \pi( (C_{high},T_{high} ) | x).
\end{equation}

Starting from a base open-source LLM $\pi$, the model is optimized to maximize the likelihood of generating these high-level thoughts conditioned on the input problems. Through this thought-based SFT, the fine-tuned LLM $\pi_{thought}$ not only learns to identify challenges in log analysis tasks but also acquires the ability to formulate structured and expert-like reasoning workflows. 

\subsection{Enhancement with Stepwise Solutions from Fine-grained}
While the first stage enables the LLM to formulate a stepwise reasoning workflow from the coarse-grained, effective log analysis also requires precise handling of details within logs.
In the second stage, we further strengthen the LLM’s analytical granularity and correctness from the fine-grained.
\subsubsection{Imitation Learning}
We first align the high-level thought with instantiated reasoning in this stage to further strengthen the LLM's ability to follow the structured workflows for instantiated
reasoning, where we utilize task-specific reasoning processes generated by the thought-enhanced LLM $\pi_{thought}$ as a source of fine-tuning data.
Specifically, we first sample instance data $D_{samples} = {(x_{sample}, y_{sample})}$ from the training sets of log analysis tasks, such as anomaly detection and root cause analysis.
Then, the LLM $\pi_{thought}$ is prompted to generate reasoning steps for each instance, following a ``think-then-answer'' paradigm, where the model not only provides problem-solving strategy step by step but also to incorporate instance-specific reasoning details within <solutions>; the final result is then provided within the <answer>. 
Finally, we filter the reasoning processes where the final answer is correct to further fine-tune the thought-enhanced LLM $\pi_{thought}$. The obtained correct dataset is denoted as:
\begin{equation}
    D_{corr} = \{(x,y,\hat{\mathbb{T}}_{traj},\hat{y})|(x,y) \in D_{samples},\hat{y} = y\},
\end{equation}
where  $\hat{\mathbb{T}}_{traj},\hat{y} = \pi_{thought}(x)$ represents the detailed solution trajectory and results generated by the $\pi_{thought}$. 

The fine-tuning process can be formulated as:
\begin{equation}
    \mathcal{L}_{solution-SFT} = argmax \sum_{(x,\hat{s}_{traj},\hat{y}) \in D_{corr}} log \pi_{thought}( (\mathbb{T}_{traj},\hat{y} ) | x).
\end{equation}

The objective is to reinforce the model to generate detailed solutions that effectively unify workflow and detail perception. After fine-tuning, we get the solution-enhanced model $\pi_{solution}$.



\subsubsection{Steps of Calibration}
Although imitation learning enhances the ability for instantiated reasoning, workflow-driven solutions may still produce reasoning steps with logical flaws, incomplete steps, or incorrect conclusions when applied to specific instances.
To correct the erroneous reasoning, we introduce preference learning that explicitly guides the LLM to discriminate between accurate and flawed reasoning steps, thereby enabling it to learn more effectively from its mistakes.
For computational efficiency, we employ DPO  as the optimization framework.

To build the preference dataset, we first identify instances from the sampled data $D_{samples}$ where the predicted answer $\hat{y}$ is inconsistent with the ground truth $y$. Then obtain the erroneous dataset:
\begin{equation}
    D_{err} = \{(x,y,\hat{\mathbb{T}}_{traj},\hat{y})|(x,y) \in D_{samples},\hat{y} \neq y\},
\end{equation}
where $\hat{\mathbb{T}}_{traj}$ represents the erroneous solution trajectory. Given that each erroneous solution is explicitly presented as a sequence of reasoning steps, it can be further represented as: 
$\hat{\mathbb{T}}_{traj}^- = \{ \{\tau_{i}^+\}_{i=1}^{k-1},\{\tau_{i}^-\}_{i=k-1}^{n} \}$, where $\tau_{i}^+$ ($\tau_{i}^-$) denotes the corrected (error) step ,$k$ denotes the first error step.

Second, to correct flawed reasoning steps, we leverage a powerful LLM as an experienced teacher model and design a prompt $\mathbb{P}_{re}$ that instructs the teacher to identify errors in the erroneous trajectory. Upon identification of errors, the teacher is prompted to reflect on the flawed step and then revise this step accordingly. Inspired by expert diagnostic practices, especially the exclusion method, the teacher explains why the flawed step is invalid, excludes it as an assumption, and reconstructs the reasoning chain from the current step. This can be denoted as:
\begin{align}
    \hat{\mathbb{T}}_{traj}^+ & = LLM(\mathbb{P}_{re},x,y,\hat{\mathbb{T}}_{traj}^-) \\
   & = \{ \{\tau_{i}^+\}_{i=1}^{k-1},\{a_{i}^+\}_{i=k-1}^{m} \},
\end{align}
where we provide the ground-truth $y$ as reference, $\hat{\mathbb{T}}_{traj}^+$ represents corrected trajectory, $a_{i}^+$ represents the revised step. Hence, we construct the preference dataset $D_{pref}$ that pairs correct and incorrect solution trajectories for each input $x$:
\begin{equation}
    D_{pref} = \{(x, \hat{\mathbb{T}}_{traj}^+, \hat{\mathbb{T}}_{traj}^-) | (x,\hat{\mathbb{T}}_{traj}^-) \in D_{err} \}.
\end{equation}

Finally, we adopt DPO to further optimize the LLM  $\pi_{solution}$ using the preference dataset.
The objective of DPO is to maximize the margin between the log-likelihood of the correct trajectory $\hat{\mathbb{T}}_{traj}^+$ and the incorrect trajectory $\hat{\mathbb{T}}_{traj}^-$, while ensuring that the updated policy remains close to the original reference policy. Hence, we define the loss function for optimizing $\pi_{solution}$ as follows:



\begin{equation}
\mathcal{L}_{solution-DPO} = -\mathbb{E}_{(x,\hat{\mathbb{T}}_{traj}^+, \hat{\mathbb{T}}_{traj}^-) \sim D_{pref}}  \\ \Biggl[ log \sigma \biggl( \beta log\frac{\pi_{dpo}(\hat{\mathbb{T}}_{traj}^+|x)}{\pi_{solution}(\hat{\mathbb{T}}_{traj}^-|x)} - \beta log\frac{\pi_{dpo}(\hat{\mathbb{T}}_{traj}^+|x)}{\pi_{solution}(\hat{\mathbb{T}}_{traj}^-|x)} \biggr) \Biggr],
\end{equation}
where $\sigma(·)$ is the sigmoid function, $\beta$ is a hyperparameter that regulates the extent of the target model’s divergence from the reference model,
$\pi_{DPO}$ represents the LLM being optimized, initialized from $\pi_{solution}$.
After preference learning, the LLM $\pi_{DPO}$ will follow the ``think-then-answer'' pattern to reason on log analysis tasks.

\section{Experiments}

\subsection{Data Preparation}
To support LogReasoner’s coarse-to-fine enhancement framework, we curate datasets from both software system logs \cite{liu2024logprompt} and network device logs \cite{ma2024knowlog}. 
For the first stage, which focuses on high-level thought construction, we collect 400 troubleshooting cases, including symptom descriptions and explanations of the flowcharts from Huawei\footnote{\url{https://support.huawei.com/enterprise/en/doc/index.html}} and H3C’s\footnote{\url{https://www.h3c.com/en/Support/Resource_Center/Technical_Documents/}} public maintenance handbooks. To further enrich the dataset, we incorporate logs from publicly available datasets. Specifically, from the LogPrompt \cite{liu2024logprompt} dataset, we utilize anomaly detection logs derived from the BGL and Spirit datasets \cite{zhu2023loghub}. LogPrompt provides 1,766 and 1,297 log templates extracted from BGL and Spirit, respectively, paired with anomaly labels. We randomly sample 150 logs (approximately 10\%) from each dataset. From the KnowLog dataset \cite{ma2024knowlog}, we sample 150 logs each from Huawei and H3C network device logs for potential failure prediction and root cause analysis.
Given that high-level thought construction with LLMs requires manual verification, the total size of the dataset for this stage is kept below 2,000 instances. 
After applying the DPP algorithm for diversity sampling, we ultimately select 1,000 diverse cases as the final dataset for thought-based SFT.

The second phase focuses on leveraging task-specific instances to enhance the details of reasoning within the established high-level framework. For this purpose, we sample a separate set of task-specific instances: 1,000 log entries each from the BGL and Spirit datasets for anomaly detection, and 1,000 log entries each from Huawei and H3C datasets for other tasks. The thought-enhanced LLM processes these samples to generate reasoning chains. Correct solutions are utilized for solution-based SFT, while incorrect reasoning chains are corrected and incorporated into DPO training. This approach ensures that the model learns both from positive examples and from calibrated corrections of its mistakes.

Detailed data statistics are provided in Table \ref{tab:data_prepare}. To prevent overfitting to any single task domain, we mixed all task datasets during training at each stage. The carefully constructed datasets and reasoning processes have been made publicly available in our repository to facilitate future research.

\begin{table}[h]
\centering
\caption{Statistics of the dataset used for reasoning enhancement.}
\label{tab:data_prepare}
\tabcolsep = 0.06cm
\resizebox{0.8\columnwidth}{!}{%
\begin{tabular}{c|ccccc}
\toprule[1.5pt]
\multicolumn{1}{l|}{Stage} & Task                         & Data Source              & Domain        & \# SFT NUM & \multicolumn{1}{l}{\# DPO NUM} \\ \midrule
\multirow{5}{*}{Stage1}    & Question Answering           & Handbook         & Huawei        & 400        & -                              \\ \cmidrule{2-6} 
                           & Anomaly Detection            & LogPrompt                & BGL \& Spirit & 300        & -                              \\ \cmidrule{2-6} 
                           & Log Semantic Matching        & \multirow{3}{*}{KnowLog} & Huawei \& H3C & 300        & -                              \\
                           & Potential Failure Prediction &                          & Huawei \& H3C & 300        & -                              \\
                           & Root Cause Analysis          &                          & Huawei        & 200        & -                              \\ \midrule
\multirow{4}{*}{Stage2}    & Anomaly Detection            & LogPrompt                & BGL \& Spirit & 1265       & 735                            \\ \cmidrule{2-6} 
                           & Log Semantic Matching        & \multirow{3}{*}{KnowLog} & Huawei \& H3C & 1626       & 374                            \\
                           & Potential Failure Prediction &                          & Huawei \& H3C & 1233       & 767                            \\
                           & Root Cause Analysis          &                          & Huawei        & 302        & 698                            \\ \bottomrule[1.5pt]
\end{tabular}%
}
\end{table}

\subsection{Implementation Details}
\subsubsection{Thought Construct Settings:}
LogReasoner relies on a powerful LLM for error correction and the construction of high-level thoughts.  In our experiments, we utilize GPT-4o as the core engine for thought generation. Specifically, we interact with GPT-4o (version: \textit{gpt-4o-2024-08-06}) via HTTP requests to the OpenAI API. To enhance the diversity of generated thoughts, the \textit{temperature} parameter of GPT-4o is set to 0.8.
For sample selection using the DPP algorithm, we employ the \textit{bge-large-en-v1.5} embedding model for semantic representation due to its superior performance among open-source embedding models.

\subsubsection{Model Tuning Setting:} \label{tuning setting}
We adopt the LLaMA-Factory framework \cite{zheng2024llamafactory} to fine-tune various instruction-following LLMs, including \textit{Qwen2.5-Instruct-1.5B} \cite{yang2024qwen2}, \textit{Qwen2.5-Instruct-7B}, and \textit{Llama3.1-Instruct-8B} \cite{grattafiori2024llama}. 
This allows us to evaluate the effectiveness of our framework across different model architectures and sizes. All training is conducted on two NVIDIA A100 GPUs.

For SFT, we use a batch size of 16, a context window of 2096 tokens, a learning rate of 3.0e-5 with a cosine learning rate scheduler, and train for 5 epochs. 
For DPO, the batch size is set to 4,  the learning rate is set to 5.0e-6, and the epoch is also set to 5. Additionally, the DPO-specific $\beta$ parameter is set to 0.1. 
All other hyperparameters follow the default settings provided by LLaMA-Factory.

\subsubsection{Evaluation Setting:} 
During inference, all open-source LLMs are hosted with vLLM\footnote{\url{https://github.com/vllm-project/vllm}}, which significantly accelerates the inference process. To ensure stability in the LLM's output, we set the temperature of the LLMs to 0 during evaluation.
For fairness in comparison, all LLMs reason based on the chain-of-thought strategy and provide the reasoning process. In addition, these LLMs follow the same in-context learning (ICL) setup, where each LLM is provided with three examples in the prompt. Additionally, the same examples are used for all LLMs to eliminate variability caused by differences in input settings.

\subsection{Evaluation Tasks}
To evaluate the reasoning capability and generalizability of LogReasoner, we evaluate it on four downstream log analysis tasks covering both software system logs and network device logs, where software log datasets are sourced from LogHub \cite{zhu2023loghub}. 
For network device logs, following KnowLog \cite{ma2024knowlog},  we collect from public product documentation and forums, focusing on Huawei and H3C with two devices: Switches and Routers. Table \ref{tab:evalution_task} provides detailed dataset statistics, and all data used during the enhancement phase is excluded from the evaluation to prevent data leakage. All datasets are made publicly available to facilitate reproducibility and future research. 

Although log parsing is a widely studied task, a recent benchmark study \cite{ma2024llmparser} shows that numerous LLMs have achieved perfect or near-perfect parsing accuracy on the public benchmark \cite{zhu2023loghub}. Furthermore, the focus of this task toward the semantic understanding of logs. Since this paper emphasizes reasoning capabilities, log parsing is excluded from our evaluation.
The four evaluation tasks and their corresponding metrics are described as follows:

\subsubsection{Anomaly Detection (AD)} Anomaly detection is one of the most widely studied tasks in log analysis \cite{zhu2023loghub,yu2024deep}, aiming to predict the presence of anomalies within log sequences. For this task, the model receives a sequence of log messages as input and determines whether an anomaly exists within the window.

\textbf{Dataset and Metric:}  We evaluate on the BGL and Spirit two datasets. To manage the computational overhead caused by the large size of the original datasets, we sample 1,000,000 log entries from each dataset and construct log sessions via fixed-window grouping (20 consecutive logs per session), resulting in 50,000 test sessions per dataset.
Following prior works on anomaly detection \cite{tao2023biglog,le2022log}, we report \textit{Precision}, \textit{Recall}, and \textit{F1-score} on the anomaly class as evaluation metrics.

\subsubsection{Log Semantic Matching (LSM)} Log Semantic Matching \cite{ma2024knowlog} is a task designed to assess whether a given log is semantically consistent with its natural language description. It evaluates the model's ability to understand the semantic relationship between logs and their textual explanations.

\textbf{Dataset and Metric:} In line with KnowLog \cite{ma2024knowlog}, we collect logs and their corresponding descriptions from public documentation. Each log is paired with its ground-truth description, forming [Log, Description] pairs. For each log, a randomly selected incorrect description is added as a negative sample.

As a binary classification task, wherein both positive and negative cases are equally important, we report \textit{Accuracy} and \textit{Weighted-F1} as evaluation metrics.

\subsubsection{Potential Failure Prediction (PFP)} This task predicts whether a given log indicates a potential system or device failure. It is framed as a binary classification problem to distinguish between failure-indicative logs and notification-indicative logs.

\textbf{Dataset and Metric:}
Logs are sourced from Huawei and H3C documentation, where logs are annotated with risk levels (e.g., Error, Informational). To avoid data leakage, we mask the risk level information in the input logs. Logs with risk levels above ``Error'' are labeled as failures; others are labeled as notifications. 
Similar to anomaly detection, we report \textit{Precision}, \textit{Recall}, and \textit{F1-score} on the failure-indicative (True) class.

\subsubsection{Root Cause Analysis (RCA)} This task involves identifying the root cause of anomalies based on log data. This task is framed as a multi-class classification problem where the model must assign logs to one of several predefined error categories.

\textbf{Dataset and Metric:}
The dataset is constructed by collecting user-submitted issues from Huawei’s public forums. Only discussions that explicitly analyzed the causes of errors are included. The final dataset contains logs categorized into five error types, such as communication failures or device failures.
Given the unbalanced nature of the dataset and the varying importance of different classes, we report \textit{Accuracy} and \textit{Weighted-F1} as evaluation metrics.

\begin{table}[]
\centering
\renewcommand{\arraystretch}{0.85}
\caption{Datasets of downstream tasks log for evaluation.}
\label{tab:evalution_task}
\resizebox{0.7\columnwidth}{!}{%
\begin{tabular}{l|cc|cc}
\toprule[1.5pt]
Tasks                        & \multicolumn{2}{c|}{BGL}       & \multicolumn{2}{c}{Spirit}    \\ \midrule
Anomaly Detection            & \multicolumn{2}{c|}{1,000,000} & \multicolumn{2}{c}{1,000,000} \\ \midrule
\multirow{2}{*}{}            & \multicolumn{2}{c|}{Huawei}    & \multicolumn{2}{c}{H3C}       \\
                             & Switches       & Routers       & Switches       & Routers      \\ \midrule
Log Semantic Matching        & 976            & 723           & 868            & 945          \\
Potential Failure Prediction & 2,563          & 2,338         & 1,508          & 2,220        \\
Root Cause Analysis          & 1,673          & 1,252         & -              & -            \\ \bottomrule[1.5pt]
\end{tabular}%
}
\end{table}

\subsection{Baselines}
Current approaches to LLM-based inference can be broadly categorized by their test-time scaling strategies into two paradigms \cite{muennighoff2025s1}: (1) Sequential reasoning, such as standard Chain-of-Thought (CoT), and (2) Parallel reasoning, such as self-consistency or majority voting \cite{wangself}. Since these reasoning strategies can be applied to various foundation LLMs, and our focus is on evaluating the reasoning capabilities of the base models themselves, we adopt the standard CoT-based in-context learning (ICL) paradigm as the primary inference method for all foundation LLMs in our comparisons. This approach is widely recognized and commonly used in LLM-based log analysis studies \cite{liu2024logprompt,xu2024divlog,jiang2023lilac}. 

To rigorously evaluate the reasoning capabilities of LogReasoner, we compare its performance against a comprehensive set of baselines spanning four distinct paradigms of language models. These categories are defined by their underlying technical approaches, and within each, we select the most representative works for comparison.

\subsubsection{Pre-trained Language Models:}
We include three typical pre-trained models as baselines: \textbf{BERT} \cite{le2021log}, \textbf{Biglog} \cite{tao2023biglog}, and \textbf{KnowLog} \cite{ma2024knowlog}. These models follow the pretrain-and-finetune paradigm, which is prevalent in log analysis before the advent of LLMs.  BERT is widely adopted as a foundational model for log representation; Biglog specializes in extracting salient features through log-specific pre-training; and KnowLog enhances log analysis performance by incorporating external knowledge during pre-training. All pre-trained models are fine-tuned on the same task-specific datasets (as indicated in Table \ref{tab:data_prepare}) to ensure fair comparison.

\subsubsection{General proprietary LLMs:}
We include two proprietary LLMs (\textbf{ChatGPT} and \textbf{GPT-4o}) as baselines due to their remarkable potential across various domains. These commercially available LLMs with strong generalization and reasoning capabilities are queried via the OpenAI API.

\subsubsection{General Open-source LLMs:}
Considering practical deployment scenarios and cost constraints, we primarily select open-source LLMs with model sizes of less than 8B parameters. 
Specifically, we evaluate \textbf{\textit{Qwen2.5-Instruct (1.5B, 3B, and 7B)}} \cite{yang2024qwen2} and \textbf{\textit{Llama3.1-Instruct-8B}} \cite{grattafiori2024llama}, covering diverse architectures and scales. In addition, we compare against \textbf{\textit{DeepSeek-R1-Distill-Qwen-7B}} \cite{guo2025deepseek}, a reasoning-enhanced LLM that distills reasoning patterns from DeepSeek.

\subsubsection{Task-Specific Chain-of-Thought Enhanced LLM:}
Constructing rationales from advanced LLMs and fine-tuning open-source LLMs on these rationales is a well-established method for improving reasoning performance on specific tasks \cite{hsieh2023distilling,kang2023knowledge}. To provide a stronger comparison for LogReasoner, we adopt this CoT-enhanced approach as a baseline. Specifically, we first employ GPT-4o to generate reasoning rationales for each downstream task via CoT prompting, and these rationales are constructed using the same input-output pairs as described in Table \ref{tab:data_prepare}.
To be fair, we also directly merge the original analytical processes collected in the first stage of LogReasoner.
Then these rationales are used to fine-tune \textit{Qwen2.5-Instruct-7B}, and \textit{Llama3.1-Instruct-8B}, following the same SFT process as described in Section \ref{tuning setting}.

\begin{table}[]
\centering
\renewcommand{\arraystretch}{0.8}
\caption{Model performance on the Anomaly Detection and Root Cause Analysis tasks. \textcolor{red}{(+X)} denotes the performance difference compared to the original foundation LLM.}
\label{tab:my-table1}
\resizebox{0.95\columnwidth}{!}{%
\begin{tabular}{lcc|cc}
\toprule[1.5pt]
\rowcolor[HTML]{FFFFFF} 
\cellcolor[HTML]{FFFFFF}                          & \multicolumn{2}{c|}{\cellcolor[HTML]{FFFFFF}Anomaly Detection (Precision / Recall / F1)}                              & \multicolumn{2}{c}{\cellcolor[HTML]{FFFFFF}RCA (Accuracy / Weighted-F1)}      \\ \cmidrule{2-5} 
\rowcolor[HTML]{FFFFFF} 
\cellcolor[HTML]{FFFFFF}                          & \multicolumn{1}{c|}{\cellcolor[HTML]{FFFFFF}}                      & \cellcolor[HTML]{FFFFFF}                         & \multicolumn{2}{c}{\cellcolor[HTML]{FFFFFF}Huawei}                            \\
\rowcolor[HTML]{FFFFFF} 
\multirow{-3}{*}{\cellcolor[HTML]{FFFFFF}Methods} & \multicolumn{1}{c|}{\multirow{-2}{*}{\cellcolor[HTML]{FFFFFF}BGL}} & \multirow{-2}{*}{\cellcolor[HTML]{FFFFFF}Spirit} & Switches                              & Routers                               \\ \midrule
\rowcolor[HTML]{FFFFFF} 
BERT                                         & \multicolumn{1}{c|}{\cellcolor[HTML]{FFFFFF}24.79 / 100.0 / 39.73} & 57.29 / 81.28 / 67.21                            & 25.88 / 27.41                         & 30.43 / 21.13                         \\
\rowcolor[HTML]{FFFFFF} 
Biglog                                            & \multicolumn{1}{c|}{\cellcolor[HTML]{FFFFFF}25.39 / 100.0 / 40.50} & 55.68 / 100.0 / 71.53                            & 26.95 / 26.60                         & 31.38 / 18.29                         \\
\rowcolor[HTML]{FFFFFF} 
KnowLog                                           & \multicolumn{1}{c|}{\cellcolor[HTML]{FFFFFF}25.31 / 90.30 / 39.53} & 57.20 / 92.49 / 70.69                            & 36.22 / 40.28                         & 38.41 / 33.36                         \\ \midrule
\rowcolor[HTML]{FFFFFF} 
ChatGPT                                           & \multicolumn{1}{c|}{\cellcolor[HTML]{FFFFFF}31.55 / 100.0 / 47.96} & 76.20 / 100.0 / 86.49                            & 9.80 / 8.92                           & 16.21 / 13.59                         \\
\rowcolor[HTML]{FFFFFF} 
GPT-4o                                            & \multicolumn{1}{c|}{\cellcolor[HTML]{FFFFFF}38.62 / 89.24 / 53.91} & 77.12 / 99.29 / 86.81                            & 35.92 / 45.69                         & 26.83 / 30.79                         \\ \midrule
\rowcolor[HTML]{FFFFFF} 
Qwen2.5-1.5B                                      & \multicolumn{1}{c|}{\cellcolor[HTML]{FFFFFF}44.72 / 18.98 / 26.65} & 78.17 / 43.49 / 55.89                            & 17.45 / 25.83                         & 8.22 / 9.35                           \\
\rowcolor[HTML]{FFFFFF} 
Qwen2.5-3B                                        & \multicolumn{1}{c|}{\cellcolor[HTML]{FFFFFF}26.17 / 94.21 / 40.97} & 61.26 / 99.53 / 75.84                            & 21.33 / 21.86                         & 28.83 / 26.14                         \\
\rowcolor[HTML]{FFFFFF} 
Qwen2.5-7B                                        & \multicolumn{1}{c|}{\cellcolor[HTML]{FFFFFF}33.06 / 99.59 / 49.64} & 64.95 / 96.51 / 77.65                            & 25.46 / 27.69                         & 29.79 / 33.83                         \\
\rowcolor[HTML]{FFFFFF} 
Llama3.1-8B                                       & \multicolumn{1}{c|}{\cellcolor[HTML]{FFFFFF}36.01 / 92.99 / 51.91} & 74.17 / 87.39 / 80.24                            & 17.63 / 22.14                         & 20.44 / 23.64                         \\
\rowcolor[HTML]{FFFFFF} 
R1-Distill-Qwen-7B                                & \multicolumn{1}{c|}{\cellcolor[HTML]{FFFFFF}30.69 / 99.10 / 46.86} & 82.32 / 79.46 / 80.87                            & 35.59 / 42.64                         & 35.22 / 36.98                         \\ \midrule
\rowcolor[HTML]{FFFFFF} 
Qwen2.5-7B-CoT-SFT                                    & \multicolumn{1}{c|}{\cellcolor[HTML]{FFFFFF}61.52 / 69.60 / 65.31} & 77.21 / 98.85 / 86.70                            & 34.48 / 40.75                         & 30.51 / 32.24                         \\
\rowcolor[HTML]{FFFFFF} 
Llama3.1-8B-CoT-SFT                                   & \multicolumn{1}{c|}{\cellcolor[HTML]{FFFFFF}61.98 / 70.17 / 65.82} & 78.71 / 98.45 / 87.48                            & 38.43 / 47.82                         & 30.03 / 33.86                         \\ \midrule
\rowcolor[HTML]{E3F8FB} 
LogReasoner (Qwen2.5-1.5B)                        & \multicolumn{1}{c|}{\cellcolor[HTML]{E3F8FB}54.96 / 74.25 / 63.17 \textbf{\redtext{(+36.5)}}} & 77.83 / 96.06 / 85.99 \textbf{\redtext{(+30.1)}}                             & 40.22 \textbf{\redtext{(+22.8)}} / 45.94 & 37.69 \textbf{\redtext{(+29.5)}} / 39.60 \\
\rowcolor[HTML]{E3F8FB} 
LogReasoner (Qwen2.5-7B)                          & \multicolumn{1}{c|}{60.87 / 95.35 / 74.31 \textbf{\redtext{(+24.7)}}} & 87.44 / 98.05 / 92.44 \textbf{\redtext{(+14.8)}}                            & 50.56 \textbf{\redtext{(+25.1)}}  / 56.79 & 39.69 \textbf{\redtext{(+9.9)}} / 41.59 \\
\rowcolor[HTML]{E3F8FB} 
LogReasoner (Llama3.1-8B)                         & \multicolumn{1}{c|}{64.75 / 94.05 / \textbf{76.70} \textbf{\redtext{(+24.8)}}} & 88.66 / 98.47 / \textbf{93.31} \textbf{\redtext{(+13.1)}}   & \textbf{54.63} \textbf{\redtext{(+37.0)}} / 58.67 & \textbf{46.92} \textbf{\redtext{(+26.5)}} / 46.32 \\ \bottomrule[1.5pt]
\end{tabular}%
}
\end{table}

\begin{table}[]
\centering
\renewcommand{\arraystretch}{0.8}
\caption{Model performance on Log Semantic Matching task. \textcolor{red}{(+X)} denotes the performance difference compared to the original foundation LLM.}
\label{tab:my-table2}
\resizebox{0.9\columnwidth}{!}{%
\begin{tabular}{lcccc}
\toprule[1.5pt]
                           & \multicolumn{4}{c}{Log Semantic Matching (Accuracy / Weighted-F1)}                         \\ \cmidrule{2-5} 
                           & \multicolumn{2}{c|}{Huawei}                                                & \multicolumn{2}{c}{H3C}       \\
\multirow{-3}{*}{Methods}  & Switches      & \multicolumn{1}{c|}{Routers}                               & Switches      & Routers       \\ \midrule
BERT                  & 75.20 / 75.19 & \multicolumn{1}{c|}{75.93 / 75.82}                         & 79.72 / 79.56 & 77.88 / 77.64 \\
Biglog                     & 80.12 / 79.93 & \multicolumn{1}{c|}{77.04 / 76.90}                         & 80.76 / 80.53 & 79.15 / 78.96 \\
KnowLog                    & 83.91 / 83.82 & \multicolumn{1}{c|}{79.11/ 79.07}                          & 83.76 / 83.72 & 81.38 / 81.30 \\ \midrule
ChatGPT                    & 72.84 / 70.91 & \multicolumn{1}{c|}{71.51 / 68.82}                         & 66.58 / 62.57 & 67.72 / 63.58 \\
GPT-4o                     & 89.44 / 89.36 & \multicolumn{1}{c|}{88.10 / 87.91}                         & 84.21 / 83.84 & 86.56 / 86.27 \\ \midrule
\rowcolor[HTML]{FFFFFF} 
Qwen2.5-1.5B               & 80.22 / 79.70 & \multicolumn{1}{c|}{\cellcolor[HTML]{FFFFFF}81.32 / 80.77} & 77.53 / 76.50 & 82.96 / 82.39 \\
\rowcolor[HTML]{FFFFFF} 
Qwen2.5-3B                 & 88.72 / 88.62 & \multicolumn{1}{c|}{\cellcolor[HTML]{FFFFFF}87.96 / 87.76} & 79.14 / 78.25 & 84.55 / 84.18 \\
\rowcolor[HTML]{FFFFFF} 
Qwen2.5-7B                 & 88.11 / 87.99 & \multicolumn{1}{c|}{\cellcolor[HTML]{FFFFFF}87.41 / 87.18} & 80.76 / 80.06 & 84.65 / 84.22 \\
Llama3.1-8B                & 94.67 / 94.67 & \multicolumn{1}{c|}{92.11 / 92.06}                         & 89.86 / 89.82 & 92.74 / 92.72 \\
R1-Distill-Qwen-7B         & 86.83 / 86.83 & \multicolumn{1}{c|}{93.77 / 93.74}                         & 82.21 / 81.98 & 85.54 / 85.43 \\ \midrule
Qwen2.5-7B-CoT-SFT             & 89.44 /89.36  & \multicolumn{1}{c|}{87.41 / 87.18}                         & 79.26 / 78.38 & 84.86 / 84.45 \\
Llama3.1-8B-CoT-SFT            & 95.95 / 95.95 & \multicolumn{1}{c|}{95.02 / 95.01}                         & 91.90 / 91.88 & 93.12 / 93.11 \\ \midrule
\rowcolor[HTML]{E3F8FB} 
LogReasoner (Qwen2.5-1.5B) & 93.44 \textbf{\redtext{(+12.2)}} / 93.43 & \multicolumn{1}{c|}{93.77 \textbf{\redtext{(+13.4)}} / 93.74} & 87.78 \textbf{\redtext{(+10.2)}} / 87.62 & 91.00 \textbf{\redtext{(+8.0)}} / 90.94 \\
\rowcolor[HTML]{E3F8FB} 
LogReasoner (Qwen2.5-7B)   & 94.26 \textbf{\redtext{(+6.2)}} / 94.25 & \multicolumn{1}{c|}{94.19 \textbf{\redtext{(+9.8)}} / 94.17} & 88.94 \textbf{\redtext{(+8.2)}} / 88.87 & 91.42 \textbf{\redtext{(+6.8)}} / 91.36 \\
\rowcolor[HTML]{E3F8FB} 
LogReasoner (Llama3.1-8B)  & \textbf{96.65} \textbf{\redtext{(+2.0)}} / 96.65 & \multicolumn{1}{c|}{\textbf{95.69} \textbf{\redtext{(+4.5)}} / 95.69} & \textbf{93.47} \textbf{\redtext{(+3.6)}} / 93.47 & \textbf{94.63} \textbf{\redtext{(+1.9)}} / 94.63 \\ \bottomrule[1.5pt]
\end{tabular}%
}
\end{table}

\subsection{Evaluations}
We evaluate LogReasoner by answering the following research questions (RQs):
\subsubsection{\textbf{RQ1: How effective is LogReasoner on log analysis tasks compared
to the current mainstream LLMs?}} 
To rigorously assess the performance of LogReasoner, we conduct extensive experiments across four distinct log analysis tasks.
The results are shown in Table \ref{tab:my-table1} - \ref{tab:my-table3}, clearly demonstrating that LogReasoner achieves substantial improvements over both its base open-source LLMs and existing state-of-the-art methods.
The superiority is particularly evident in complex, reasoning-intensive tasks such as root cause analysis (RCA), where LogReasoner boosts the accuracy of Llama3.1-8B by an average of 26.24\%.
Across all tasks and datasets, LogReasoner (based on Llama3.1-8B) consistently delivers the best performance, validating the effectiveness of our reasoning enhancement framework.

\begin{table*}[]
\centering
\renewcommand{\arraystretch}{0.83}
\caption{Model performance on the Potential Failure Prediction task. \textcolor{red}{(+X)} denotes the performance difference compared to the original foundation LLM.}
\label{tab:my-table3}
\resizebox{\columnwidth}{!}{%
\begin{tabular}{lcccc}
\toprule[1.5pt]
                           & \multicolumn{4}{c}{Potential Failure Prediction (Precision / Recall / F1)}                                                                 \\ \cmidrule{2-5} 
                           & \multicolumn{2}{c|}{Huawei}                                                                & \multicolumn{2}{c}{H3C}                       \\
\multirow{-3}{*}{Methods}  & Switches              & \multicolumn{1}{c|}{Routers}                                       & Switches              & Routers               \\ \midrule
BERT                  & 69.10 / 76.09 / 72.43 & \multicolumn{1}{c|}{87.35 / 62.68 / 72.98}                         & 73.84 / 70.86 / 72.32 & 66.85 / 58.29 / 62.27 \\
Biglog                     & 67.64 / 82.72 / 74.42 & \multicolumn{1}{c|}{86.33 / 66.57 / 75.17}                         & 70.85 / 82.35 / 76.17 & 60.24 / 70.28 / 64.88 \\
KnowLog                    & 66.87 / 85.17 / 74.92 & \multicolumn{1}{c|}{85.31 / 69.72 / 76.73}                         & 72.01 / 80.94 / 76.22 & 64.15 / 76.65 / 69.84 \\ \midrule
ChatGPT                    & 67.51 / 79.79 / 73.14 & \multicolumn{1}{c|}{76.73 / 77.85 / 77.28}                         & 73.50 / 74.39 / 73.94 & 64.75 / 74.61 / 69.33 \\
GPT-4o                     & 72.13 / 80.15 / 75.93 & \multicolumn{1}{c|}{80.12 / 80.21 / 80.17}                         & 73.55 / 74.59 / 74.07 & 67.18 / 75.84 / 71.25 \\ \midrule
\rowcolor[HTML]{FFFFFF} 
Qwen2.5-1.5B               & 73.05 / 55.58 / 63.13 & \multicolumn{1}{c|}{\cellcolor[HTML]{FFFFFF}85.48 / 50.45 /63.45}  & 73.65 / 51.05 / 60.31 & 64.46 / 62.12 / 64.19 \\
\rowcolor[HTML]{FFFFFF} 
Qwen2.5-3B                 & 70.73 / 64.81 / 67.64 & \multicolumn{1}{c|}{\cellcolor[HTML]{FFFFFF}82.44 / 57.29 / 67.60} & 73.34 / 60.00 / 66.00 & 66.40 / 62.12 / 64.19 \\
\rowcolor[HTML]{FFFFFF} 
Qwen2.5-7B                 & 70.51 / 62.84 / 66.45 & \multicolumn{1}{c|}{\cellcolor[HTML]{FFFFFF}82.85 / 58.28 / 68.43} & 77.69 / 50.90 / 61.51 & 68.07 / 60.95 / 64.31 \\
Llama3.1-8B                & 74.64 / 69.33 / 71.89 & \multicolumn{1}{c|}{83.37 / 79.42 / 81.35}                         & 73.77 / 77.41 / 75.55 & 66.50 / 78.18 / 71.87 \\
R1-Distill-Qwen-7B         & 70.67 / 77.99 / 74.15 & \multicolumn{1}{c|}{83.92 / 76.83 / 80.22}                         & 75.75 / 80.94 / 78.26 & 66.57 / 79.54 / 72.48 \\ \midrule
Qwen2.5-7B-CoT-SFT             & 72.78 / 72.74 / 72.76 & \multicolumn{1}{c|}{84.87 / 71.47 / 77.60}                         & 73.68 / 64.91 / 69.02 & 66.93 / 67.63 / 67.28 \\
Llama3.1-8B-CoT-SFT            & 70.45 / 80.09 / 74.96 & \multicolumn{1}{c|}{84.08 / 77.73 / 80.78}                         & 72.88 / 78.32 / 75.51 & 66.09 / 83.17 / 73.65 \\ \midrule
\rowcolor[HTML]{E3F8FB} 
LogReasoner (Qwen2.5-1.5B) & 71.68 / 78.24 / 74.82 \textbf{\redtext{(+11.7)}} & \multicolumn{1}{c|}{83.86 / 75.59 / 79.51 \textbf{\redtext{(+16.1)}}} & 72.78 / 74.39 / 73.57 \textbf{\redtext{(+11.7)}} & 65.52 / 77.29 / 70.92 \textbf{\redtext{(+14.4)}} \\
\rowcolor[HTML]{E3F8FB} 
LogReasoner (Qwen2.5-7B)   & 69.35 / 83.32 / 75.69 \textbf{\redtext{(+9.2)}} & \multicolumn{1}{c|}{82.22 / 82.63 / 82.42 \textbf{\redtext{(+14.0)}}} & 73.58 / 80.04 / 76.67 \textbf{\redtext{(+15.2)}} & 67.12 / 81.88 / 73.77 \textbf{\redtext{(+9.5)}} \\
\rowcolor[HTML]{E3F8FB} 
LogReasoner (Llama3.1-8B)  & 69.85 / 88.62 / \textbf{78.13} \textbf{\redtext{(+6.2)}} & \multicolumn{1}{c|}{80.28 / 89.21 / \textbf{84.51} \textbf{\redtext{(+5.3)}}} & 71.42 / 92.92 / \textbf{80.77} \textbf{\redtext{(+5.2)}} & 64.55 / 93.54 / \textbf{76.39} \textbf{\redtext{(+4.5)}} \\ \bottomrule[1.5pt]
\end{tabular}%
}
\end{table*}

Specifically, When compared to representative pre-trained models, including BERT, BigLog, and KnowLog, LogReasoner achieves remarkable performance gains. 
On the BGL dataset, for instance, it improves the F1-score by 36.97\%, 36.2\%, and 37.17\% over these baselines, respectively. More importantly, while conventional methods typically depend heavily on large-scale labeled data and task-specific fine-tuning, LogReasoner effectively enhances reasoning capability with minimal annotation overhead.
Furthermore, unlike black-box pre-trained models that only provide final predictions, LogReasoner generates explicit, human-readable reasoning chains, greatly improving interpretability and facilitating trust in real-world operational environments.

LogReasoner also exhibits strong competitiveness when evaluated against powerful commercial LLMs such as ChatGPT and GPT-4o, with the performance gap being particularly remarkable on complex log analysis tasks. For instance, in the root cause analysis task, LogReasoner (Llama3.1-8B) achieves accuracy advantages of 37.77\% and 19.4\% over ChatGPT and GPT-4o, respectively. These results reveal that the reasoning capabilities of commercial LLMs in log analysis remain limited, and significant adaptation gaps persist when applying general-purpose AI engines to log analysis scenarios. Additionally, LogReasoner avoids the high API costs and potential privacy concerns associated with commercial models, making it more practical for real-world deployment.

When compared to their original open-source LLMs, the reasoning capabilities of foundation LLMs are significantly enhanced by LogReasoner.   For example, in anomaly detection, LogReasoner achieves average F1-score improvements of 33.31\% on Qwen-1.5B, 19.73\% on Qwen-7B, and 18.93\% on Llama-8B.  Notably, the improvement is more pronounced for smaller models, which typically exhibit weaker baseline reasoning capabilities, underscoring the general applicability and scalability of our framework. LogReasoner is model-agnostic and can be effectively applied to diverse LLM architectures, highlighting its flexibility.

Even when compared to CoT-enhanced methods, LogReasoner maintains a clear advantage. On both anomaly detection and root cause analysis tasks, LogReasoner (Llama3.1-8B) exceeds Llama3.1-8B-CoT by 8.35\% in average F1-score and 16.53\% in accuracy. This indicates that task-specific CoT distillation is insufficient to elicit strong reasoning in log analysis, where models tend to mimic reasoning chains superficially without indeed comprehension and cannot reflect on and learn from their own mistakes. In contrast, LogReasoner incorporates a deeper thinking framework that enables it to analyze logs with greater precision and adaptability.

\begin{tcolorbox}[
    colback=gray!5!white,
    colframe=gray!95!black,
    fonttitle=\bfseries,
    arc=2.8pt,
    boxrule=0.9pt,
    enhanced,
    boxsep=1pt,              
    left=2.2pt, right=2.2pt,     
    top=2.2pt, bottom=2.2pt,     
    before upper={\setstretch{0.9}}
]
\textbf{Conclusion 1}: 
Coarse-to-fine reasoning enhancement significantly improves the reasoning of LLMs on log analysis tasks, bridging the gap between general-purpose LLMs and domain-specific log analysis. 
In addition, LogReasoner can be flexibly integrated with different LLM architectures and scales, facilitating practical deployment in automated log analysis.
\end{tcolorbox}


\begin{figure}[]
  \centering
  \includegraphics[width=0.9\columnwidth]{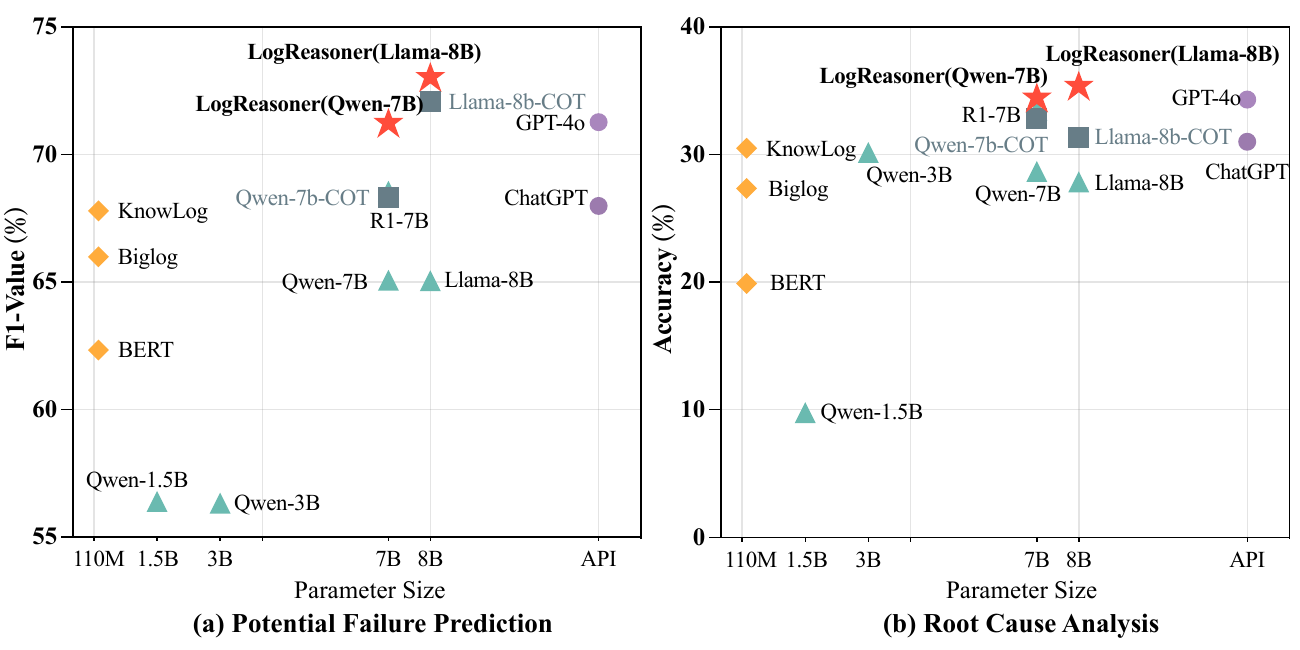}
  \caption{Out-of-distribution Performance of LogReasoner.}
  \label{fig:logreason_ood}
\end{figure}

\subsubsection{\textbf{RQ2: How effective is LogReasoner in generalization
ability?}}
To systematically evaluate the generalization capability of LogReasoner, we design a rigorous cross-device evaluation protocol that tests its performance on completely unseen types of log data. During training, LogReasoner's reasoning datasets are constructed exclusively using log sources from Switches and Routers, representing a specific operational domain. To challenge its generalization capacity, we collect evaluation datasets from Security devices, which exhibit different log structures, event types, and failure modes not encountered during training. This setup provides a robust out-of-distribution (OOD) assessment, simulating real-world scenarios where models must handle logs from new system types.
This RQ focuses on two relatively complex tasks: potential failure prediction and root cause analysis. For the potential failure prediction task, we report the average F1-score on the Huawei and H3C-Security datasets. For RCA, we only report accuracy on Huawei-Security due to the lack of H3C data.

As illustrated in Figure \ref{fig:logreason_ood}, LogReasoner demonstrates superior generalization performance, achieving the best results across both tasks. Specifically, on the potential failure prediction task, LogReasoner (Llama-8B) attains an average F1-score of 73.04\%, outperforming GPT-4o by 2\%. In root cause analysis, LogReasoner achieves an accuracy of 35.4\%, exceeding GPT-4o by 1.1\%. These results are particularly noteworthy given that GPT-4o is a large-scale commercial model with broad pre-training coverage, yet LogReasoner generalizes more effectively to unfamiliar log domains.

Compared to the original foundation LLM (Llama-8B), LogReasoner demonstrates notable improvements in both tasks, boosting F1-score and accuracy by 7.99\% and 7.54\%, respectively. 
This highlights that our approach not only enhances in-distribution performance but also enables the model to better analyze unseen data by improving both high-level and detailed reasoning.
Furthermore, LogReasoner surpasses models enhanced with task-specific CoT prompting, indicating that its advantages stem not from superficial pattern imitation but from a deeper, more adaptable form of reasoning. By instilling expert-like analytical workflows and supporting them with calibrated step-wise reasoning, LogReasoner enables more robust and context-sensitive understanding, making it particularly effective for log analysis in diverse and evolving operational environments.

\begin{tcolorbox}[
    colback=gray!5!white,
    colframe=gray!95!black,
    fonttitle=\bfseries,
    arc=2.8pt,
    boxrule=0.9pt,
    enhanced,
    boxsep=1pt,              
    left=2.2pt, right=2.2pt,     
    top=2.2pt, bottom=2.2pt,     
    before upper={\setstretch{0.9}}
]
\textbf{Conclusion 2}: 
LogReasoner achieves state-of-the-art performance on unseen log data from different devices, demonstrating strong generalization ability. Our framework can serve as a robust and adaptable solution for real-world log analysis tasks, where logs often vary across devices and environments.
\end{tcolorbox}

\subsubsection{\textbf{RQ3: How effective is LogReasoner with different proportions of reasoning data?}}
LogReasoner requires data for learning in both stages, particularly for constructing high-level thought in the first stage, which involves manual collection and validation. To investigate the impact of data quantity on LogReasoner's performance, we conduct experiments by sampling different proportions of data for both stages and evaluate the model on potential failure prediction and root cause analysis tasks using Llama3-8B as the base model.
For the first stage, we sample subsets of the collected 1,000 high-level thought data in increments of 200 and directly evaluate only with the thought-enhanced model $\pi_{thought}$. For the second stage, we sample subsets of task-specific data from the datasets in Table \ref{tab:data_prepare} at 20\% increments, and further train the model initialized from the 1000-sample thought-enhanced checkpoint.

\begin{figure}[]
  \centering
  \includegraphics[width=0.9\columnwidth]{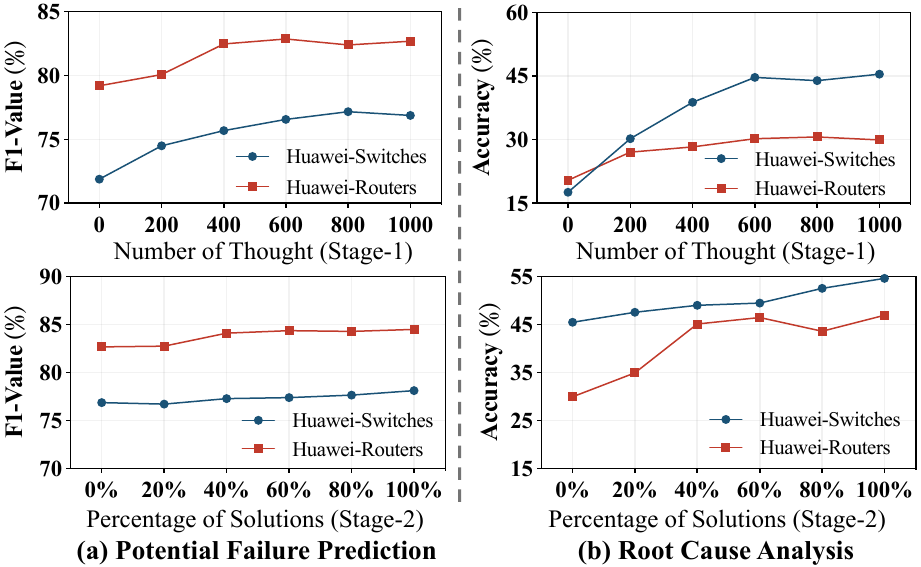}
  \caption{Performance of LogReasoner with different proportions
of reasoning data.}
  \label{fig:logreason_percentage}
\end{figure}

The results are presented in Figure \ref{fig:logreason_percentage}, which demonstrates that the LLM's performance improves with increased data size in both stages. Specifically, for the first stage, the model's performance increases sharply with the addition of diverse high-level thought data, but the improvements plateau after reaching 600 samples. 
For example, in the RCA task, LogReasoner achieves an average accuracy improvement of 18.46\% over the original base model with 600 samples, while increasing the sample size to 1,000 results in only a marginal gain of 0.21\% in accuracy. This indicates that the diversity of reasoning data is more critical than volume, and the workload for manual verification remains manageable.

In the second stage, the model's performance improvement slows after using 40\% of the task-specific data, highlighting that a relatively small amount of annotated data is sufficient for effective fine-tuning. However, we observe that for the more complex RCA task, the model remains sensitive to additional data, with performance continuing to improve as the data size increases. This suggests that complex log analysis tasks, which involve intricate reasoning, benefit significantly from larger datasets due to the higher likelihood of error correction through reflection and reasoning refinement.

\begin{tcolorbox}[
    colback=gray!5!white,
    colframe=gray!95!black,
    fonttitle=\bfseries,
    arc=2.8pt,
    boxrule=0.9pt,
    enhanced,
    boxsep=1pt,              
    left=2.2pt, right=2.2pt,     
    top=2.2pt, bottom=2.2pt,     
    before upper={\setstretch{0.9}}
]
\textbf{Conclusion 3}: 
LogReasoner can achieve competitive performance with limited high-level thought and task-specific data. 
High-level thought relies more on data diversity than volume, and task-specific detail enhancement for complex tasks benefits from larger datasets due to their higher reasoning complexity. 
\end{tcolorbox}
\vspace{-3mm}

\subsubsection{\textbf{RQ4: How effective is LogReasoner in constructing corrected solutions with different LLMs?}}
In the second stage of LogReasoner, constructing detailed step-by-step solutions is critical for improving the details of reasoning.  To achieve this, we employ GPT-4o as a teacher model to reflect on and refine the reasoning steps of the thought-enhanced open-source LLM $\pi_{thought}$ from the first stage. This approach leverages preference learning to correct reasoning errors and improve solution quality.
However, given practical constraints such as API costs and data privacy in real-world industrial settings, we also investigate an alternative self-correction mechanism. In this setup, open-source LLMs, like Qwen-7B and Llama-8B ($\pi_{thought}$), autonomously reflect on and refine their own reasoning steps without external supervision. This allows us to assess whether competitive correction quality can be achieved using only open-source resources.
We focus our evaluation on two relatively complex tasks: potential failure prediction and root cause analysis.

\begin{figure}[]
  \centering
  \includegraphics[width=0.88\columnwidth]{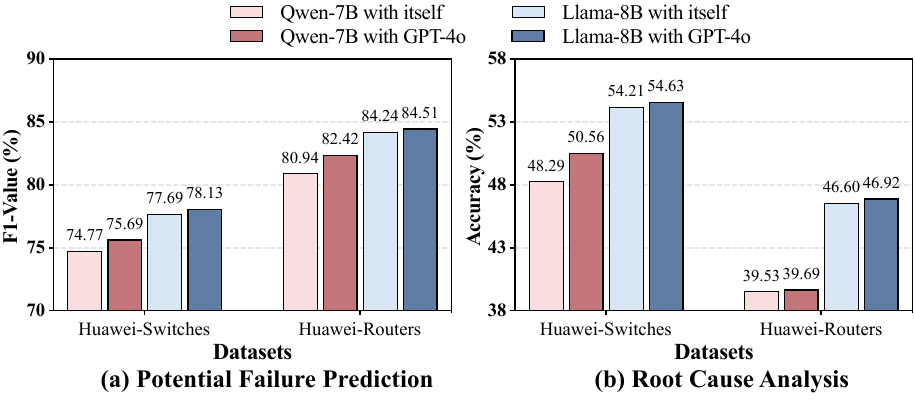}
  \caption{Performance of LogReasoner with self-generated preference data.}
  \label{fig:logreason_selfdpo}
\end{figure}

As illustrated in Figure \ref{fig:logreason_selfdpo}, models refined using GPT-4o as the teacher achieve the highest performance across tasks. For potential failure prediction, the average F1 scores of self-corrected Qwen-7B and Llama-8B are 1.2\% and 0.36\% lower, respectively, than the GPT-4o-guided model. Similarly, in RCA, self-correction leads to accuracy reductions of 1.06\% (Qwen-7B) and 0.37\% (Llama-8B) compared to the GPT-4o model.
Despite the slight advantage of GPT-4o-assisted refinement, the self-correction approach still delivers strong and competitive results. This indicates that open-source LLMs can still achieve competitive performance through self-generated corrected solutions. Such a finding has important practical implications: it confirms that organizations can deploy high-performance log analysis systems using fully local, open-source models without compromising significantly on effectiveness, enabling broader adoption under privacy-sensitive or budget-aware conditions.

\begin{tcolorbox}[
    colback=gray!5!white,
    colframe=gray!95!black,
    fonttitle=\bfseries,
    arc=2.8pt,
    boxrule=0.9pt,
    enhanced,
    boxsep=1pt,              
    left=2.2pt, right=2.2pt,     
    top=2.2pt, bottom=2.2pt,     
    before upper={\setstretch{0.9}}
]
\textbf{Conclusion 4}: 
LogReasoner demonstrates robust effectiveness in constructing corrected solutions with both commercial and open-source LLMs.  Although GPT-4o offers superior guidance, open-source LLMs are capable of self-correction with minimal performance loss, providing a practical alternative.
\end{tcolorbox}


%


\section{Ablation Studies}

In this section, we conduct ablation studies on LogReasoner to analyze two aspects: (1) the effectiveness of intermediate reasoning processes, and (2) the contributions of different enhancement stages to log analysis performance. The ablation experiments are conducted on the potential failure prediction and root cause analysis tasks using Llama3-8B as the foundation model.

\begin{figure}[]
  \centering
  \includegraphics[width=0.88\columnwidth]{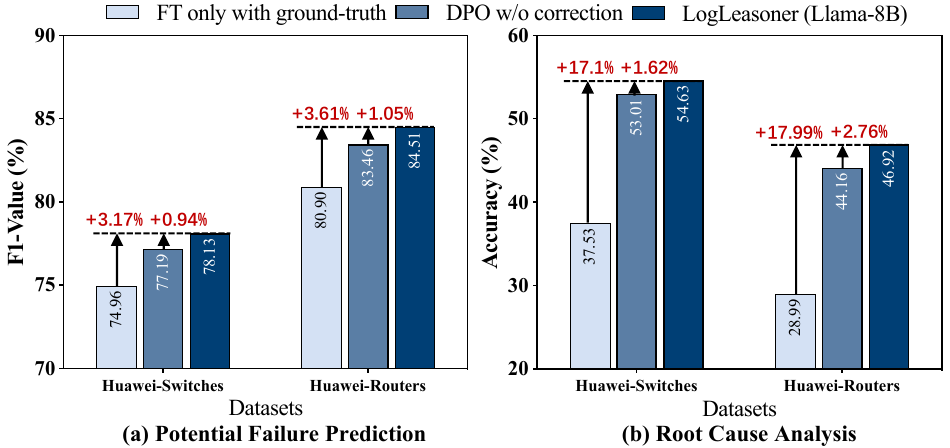}
  \caption{Ablation of intermediate reasoning processes.}
  \label{fig:logreason_process}
\end{figure}

\subsection{Ablation on Intermediate Steps}
LogReasoner incorporates intermediate reasoning steps at both the SFT and DPO stages. To evaluate their impact, we compare two ablation settings: (1) Direct Fine-tuning: we fine-tune the LLM only with the (input, ground truth) pairs from Table \ref{tab:data_prepare}, omitting all intermediate reasoning details.
(2) DPO without correction:  During preference learning, we only use ground-truth labels as correct trajectories, without involving teacher-corrected reasoning chains.

As shown in Figure \ref{fig:logreason_process}, removing explicit intermediate steps leads to a substantial performance drop, particularly on more complex log analysis tasks. Specifically, eliminating intermediate steps during STF results in a significant performance decline, with the average accuracy on RCA dropping 17.54\%.
In the DPO stage, removing the correction process and relying solely on ground truth for optimization also leads to degraded performance.
These findings underscore the crucial role of explicit intermediate reasoning in enhancing the LLM's reasoning capabilities for log analysis tasks.

\begin{figure}[]
  \centering
  \includegraphics[width=0.88\columnwidth]{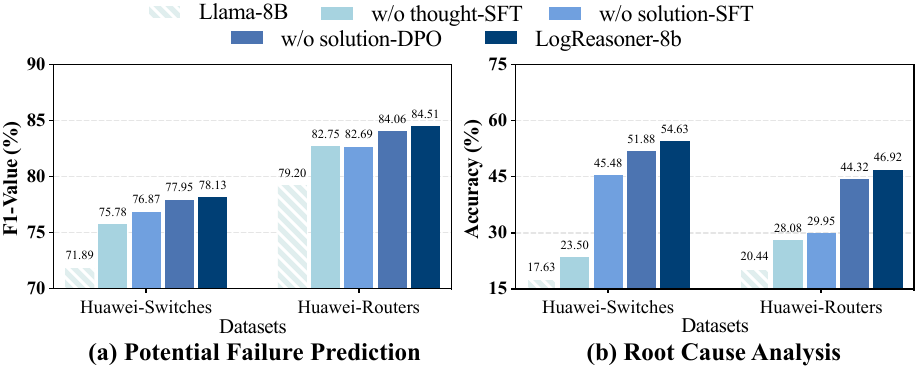}
  \caption{Ablation on different enhancement stages.}
  \label{fig:logreason_ablation}
\end{figure}

\subsection{Ablation on Enhancement Stages}
To evaluate the effectiveness of different stages in LogReasoner, we conduct experiments by removing each training objective and analyzing its impact.
The result in Figure \ref{fig:logreason_ablation} shows that the LogReasoner achieves the best performance when all training objectives are included, while omitting any stage leads to varying degrees of degradation. Specifically, for the first stage, removing the high-level thought enhancement leads to the most significant drop in performance. For example, the average accuracy on the RCA task decreased by 25.99\%. 
This suggests that the LLM struggles to develop effective reasoning abilities without high-level thought enhancement, which is critical for teaching the LLM how to analyze and reason systematically.
For the second stage, comparing the importance of the two objectives in this stage, we find that removing solution-SFT causes a more significant performance drop. When the model directly uses DPO without solution-SFT, the decline is pronounced. This indicates that solution-SFT plays a crucial role in helping the model understand how to instantiate reasoning effectively, providing a solid foundation for further preference learning.

\section{Discussion}
\subsection{Practicality of LogReasoner}
LogReasoner is designed not only to enhance the reasoning capabilities of LLMs in log analysis but also to serve as a practical and deployable solution in real-world environments. 

\textbf{Scalability.} LogReasoner exhibits strong scalability, making it suitable for organizations with varying data resources and computational constraints. As shown in RQ3, the framework achieves notable performance improvements even when trained on limited labeled log data. This data efficiency is particularly valuable in real-world settings, where acquiring large-scale, high-quality annotated logs is often challenging and expensive. Moreover, LogReasoner is model-agnostic by design. It can be effectively applied to a range of open-source backbone LLMs, from smaller models like Qwen-1.5B to larger ones such as Llama-8B, without requiring fundamental architectural changes. This flexibility ensures that LogReasoner can continually evolve and benefit from advances in foundational LLMs.

\textbf{Cost Efficiency.}
LogReasoner provides a cost-effective alternative to reliance on commercial LLM APIs, balancing performance with operational affordability. The framework supports open-source models of varying sizes, enabling users to select an appropriate model based on their computational budget and latency requirements. For instance, smaller models like Qwen-1.5B can be deployed in resource-constrained settings while still achieving significant performance gains through LogReasoner’s enhancement pipeline. Experimental results confirm that open-source models enhanced with LogReasoner can outperform commercial counterparts such as GPT-4o, thereby eliminating recurring API costs and mitigating data privacy risks associated with third-party services.

Although LogReasoner leverages GPT-4o during the data construction phase, the associated expense is modest and one-time. Specifically, generating a single high-level thought costs approximately \textbf{\$0.015}, and constructing one piece of preference data averages \textbf{\$0.018}. With typical dataset sizes, the total data preparation cost remains under \$100, and this low entry barrier makes LogReasoner accessible to small-sized enterprises.

\subsection{Case Analysis}
\subsubsection{Reasoning Case}
As shown in Fig. \ref{fig:reason_examples},  we compare the reasoning processes and conclusions of the Llama3-8B model, the powerful GPT-4o, and our proposed LogReasoner framework. 
Both Llama3-8B and GPT-4o incorrectly classify the log as "abnormal", but for different reasons. Llama3-8B employs a simplistic, keyword-triggered reasoning process, concluding an anomaly primarily based on the phrase "failed password". In contrast, GPT-4o produces a more structured yet critically flawed analysis. It misinterprets standard log anonymization placeholders as deliberate obfuscation attempts and incorrectly flags the use of a non-standard port (3564) as inherently suspicious. These errors stem from a lack of domain-specific knowledge and an inability to contextualize log events beyond superficial patterns.

LogReasoner, however, demonstrates a robust, multi-step reasoning process akin to expert troubleshooting. First, it decomposes the log entry into key components, establishing a clear basis for analysis. Second, it evaluates conflicting hypotheses, acknowledging that a failed login could indicate a security event but correctly weighing this against the operational context, where such events are common and often normal in publicly accessible systems. Most importantly, it exhibits a self-reflective capability to pause and reassess its initial impression ("Wait a second, let’s ensure this is right") before coming to a final, correct judgment. This case underscores that LogReasoner elevates LLM reasoning beyond simple pattern matching to context-aware, logical analysis. 


\subsubsection{High-level Thought Example}

\begin{figure}[]
  \centering
  \includegraphics[width=\columnwidth]{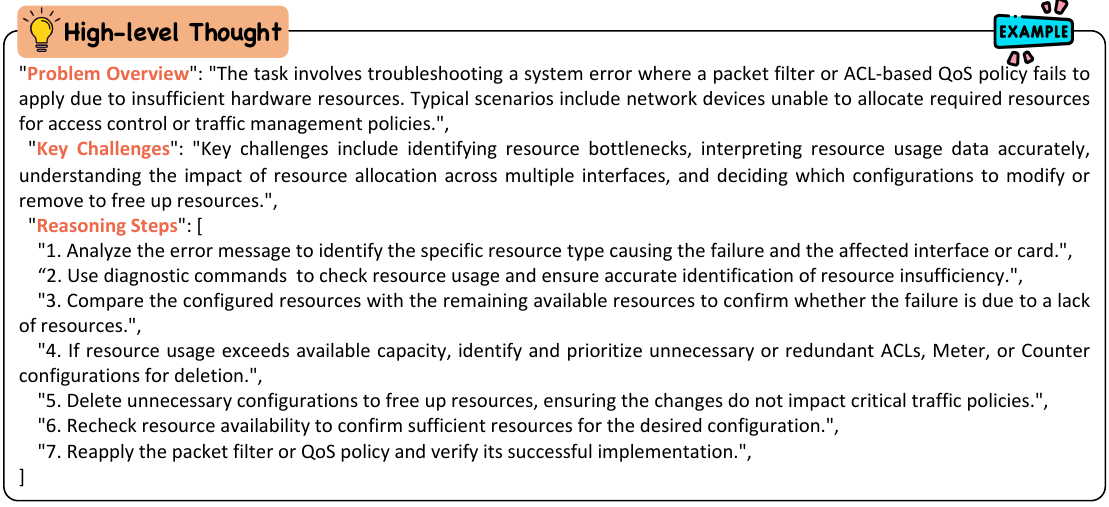}
  \caption{Example of high-level thought constructed with collected troubleshooting flowchart.}
  \label{fig:logreason_expert}
\end{figure}

To facilitate an intuitive understanding of the constructed high-level thought, we provide a specific example in Figures \ref{fig:logreason_expert}.
The high-level thought captures the following elements:
\begin{itemize}
    \item \textbf{Problem Overview:} A clear statement of the log analysis problem.
    \item \textbf{Key Challenges:} Identification of the critical challenges encountered during the analysis.
    \item \textbf{Reasoning Steps:} Structured reasoning workflow generalized from the original analysis without including instance details.
\end{itemize}
This abstraction is designed to organize reusable reasoning logic and workflows derived from expert experience, making it applicable to similar log analysis scenarios.

\begin{figure}[]
  \centering
  \includegraphics[width=0.88\columnwidth]{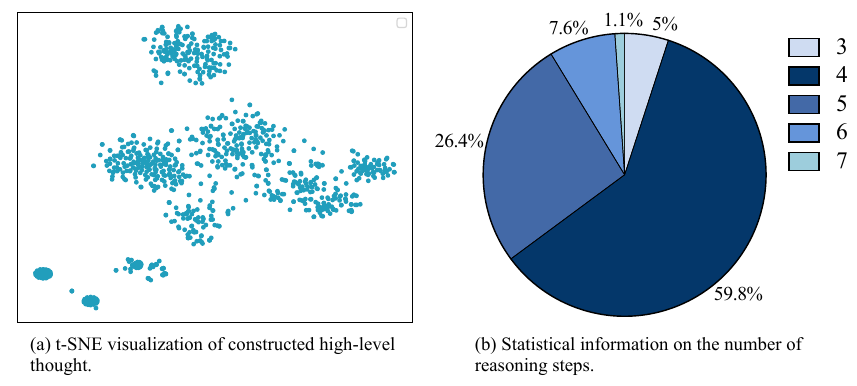}
  \caption{Statistical analysis of constructed high-level thought.}
  \label{fig:thought}
\end{figure}

To further analyze the reasoning patterns of collected high-level expert thoughts, we visualize the distribution of these high-level thoughts with t-SNE, as shown in Figure~\ref{fig:thought} (a). The visualization reveals that the semantic representations of these high-level thoughts are relatively scattered, indicating the richness of the constructed expert thinking.
In addition, we examine the number of steps in high-level thought. Figure~\ref{fig:thought} (b) shows a strong concentration, with the majority of reasoning steps consisting of 4 to 5 steps, accounting for over 86\% of all cases.

\begin{figure}[]
  \centering
  \includegraphics[width=\textwidth]{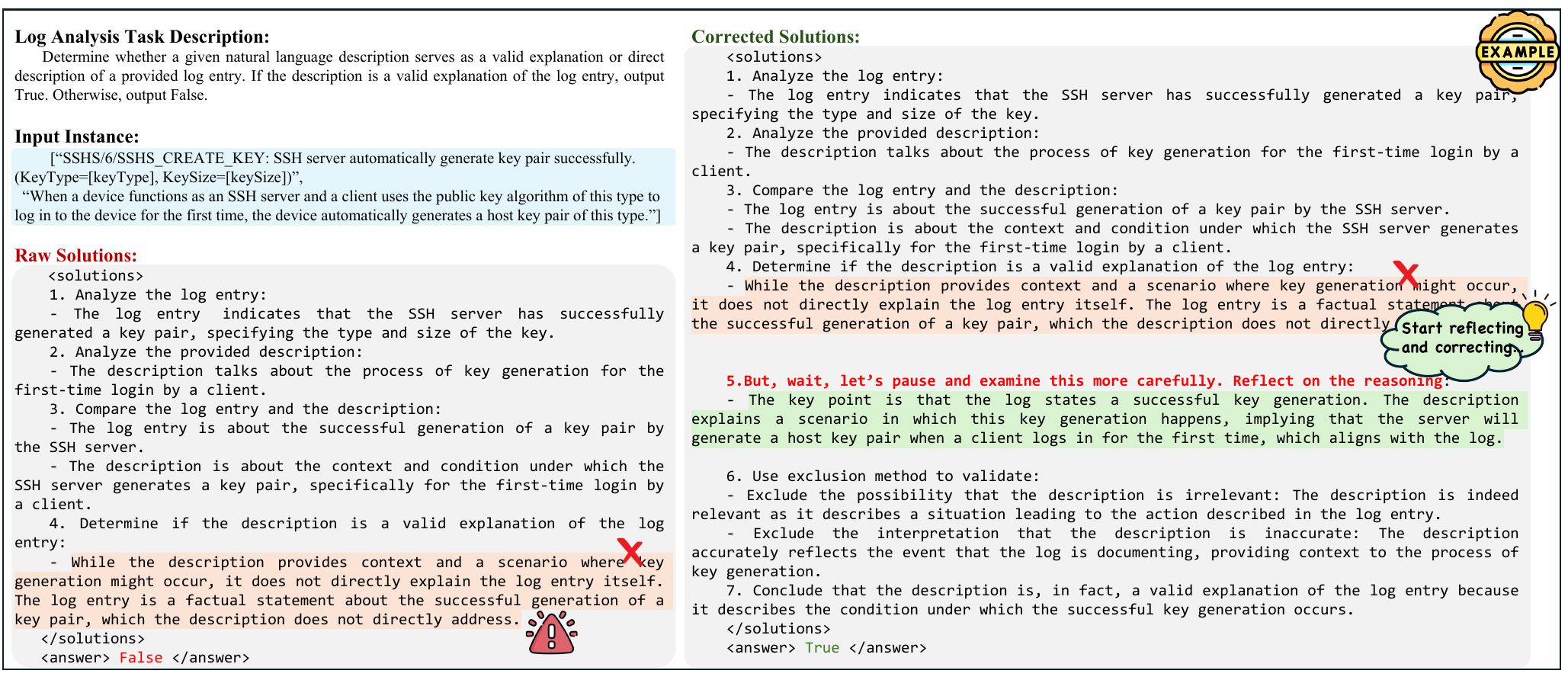}
  \caption{Example of error correction with GPT-4o.}
  \label{fig:logreason_example2}
\end{figure}

\subsubsection{Example of Corrected Data}
This section provides an example of the process of error correction with GPT-4o.
As shown in Fig. \ref{fig:logreason_example2}, the example demonstrates how our framework leverages GPT-4o to identify and correct flawed reasoning trajectories generated by the thought-enhanced model.  The key mechanism is the systematic application of an exclusion method: by analyzing the point of failure in an incorrect reasoning chain, the teacher model (GPT-4o) reflects on the erroneous assumption, excludes it, and reconstructs a logically sound path forward from that step. This process ensures that the preference data used for DPO training is accurate.

The example showcases this corrective workflow: We first collect the flawed reasoning from $\pi_{thought}$. Then, GPT-4o pinpoints the exact step where the reasoning deviates, articulating why the specific assumption is incorrect or lacks contextual support. At the same time, the teacher model formally discards the faulty assumption and rebuilds the reasoning sequence from that juncture, adhering to correct domain knowledge and logical principles. Finally, we take a revised, accurate reasoning chain as the preferred solution in the DPO pair.

\subsection{Threats to Vadility}
Despite the promising results achieved by LogReasoner, several limitations may affect the performance of this framework, which could guide future research directions.

\textit{(1) Hallucinations in LLMs.}  While it demonstrates competitive reasoning capabilities on log analysis tasks,  LogReasoner occasionally exhibits hallucinations in its generated outputs. This issue arises primarily from the lack of log-specific knowledge, which may result in factual errors.
Hallucinations \cite{huang2025survey} are a well-known challenge in LLMs and have been a focus of attention in AI research. To mitigate this limitation, future work could enhance LogReasoner with a retrieval module \cite{gu2025effectiveness} to provide relevant background knowledge and reduce factual errors.

\textit{(2) Randomness in LLM Inference.} The inherent randomness in LLM inference introduces variability in outputs, even for identical inputs. This randomness can influence the model's performance and affect the reproducibility of results.  To minimize this risk, we set the LLM's \textit{temperature} parameter to 0 during inference. This adjustment ensures deterministic outputs for the same input logs, minimizing the impact of randomness on our experimental results.

\textit{(3) Limited Experimental Scope.} LogReasoner has been validated on a limited set of datasets, which may not fully capture the diversity of real-world operational scenarios. Given the varying environments and requirements across different maintenance and operational contexts, this limitation poses a potential external threat to the generalizability of the model.
Expanding the evaluation to include a broader range of datasets and deploying LogReasoner in more realistic, real-world scenarios would help enhance its robustness and applicability.

\section{Conclusion}
In this paper, we present LogReasoner, a coarse-to-fine reasoning enhancement framework designed to address the limitations of general-purpose LLMs in log analysis. By the two-stage enhancement, LogReasoner equips LLMs with log-specific reasoning capabilities. The coarse-grained enhancement enables LLMs to formulate expert-like reasoning workflows, and the fine-grained enhancement ensures precision from the details of reasoning steps.
Our extensive experiments across four distinct log analysis tasks, using open-source LLMs such as Qwen-2.5 and Llama-3, demonstrate the effectiveness of LogReasoner.
Notably, LogReasoner significantly enhances the reasoning performance of LLMs, achieving substantial improvements in various log analysis tasks.
Ablation analysis demonstrates the effectiveness of each stage for enhancing the reasoning capabilities to analyze logs.

\section{Data Availability}
Our source
code, detailed experimental data, and results are available at \url{https://github.com/LeaperOvO/LogReasoner}.

\bibliographystyle{ACM-Reference-Format}
\bibliography{ref}

\appendix

\section{Prompt Template}

\begin{figure}[h]
  \centering
  \includegraphics[width=\columnwidth]{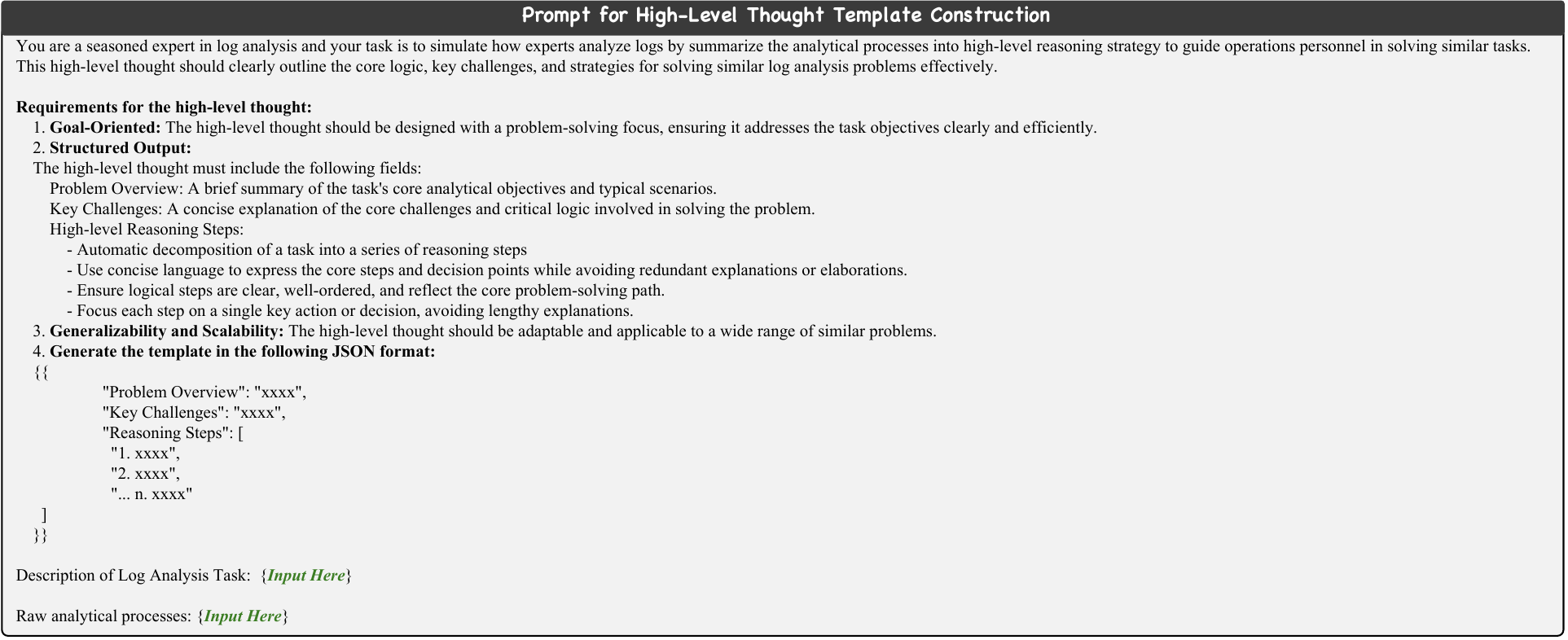}
  \caption{Prompt template for high-level thought construction.}
  \label{fig:logreason_thought_template}
\end{figure}

Figure \ref{fig:logreason_thought_template} shows the prompt template for high-level thought construction. This prompt is designed to distill raw analytical processes into structured, reusable high-level thought templates for log analysis tasks. It encourages the generation of high-level thoughts with a clear focus on solving specific log analysis problems. 
The prompt includes identifying core challenges and outlining concise and scalable problem-solving steps. The output, structured in JSON format, ensures clarity and adaptability.

\begin{figure}[]
  \centering
  \includegraphics[width=\columnwidth]{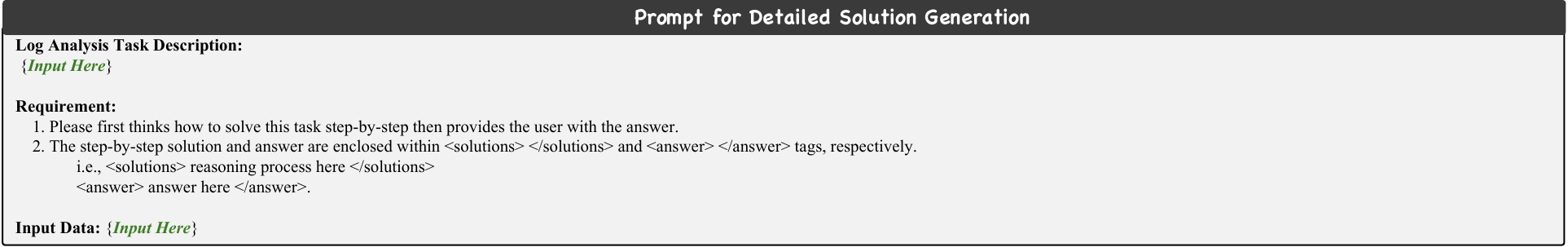}
  \caption{Prompt template for detailed solution generation.}
  \label{fig:logreason_solution_prompt}
\end{figure}

Figure \ref{fig:logreason_solution_prompt} shows the prompt template for detailed solution generation in the second stage with the thought-enhanced LLM $\pi_{thought}$.
This prompt is designed to guide the generation of detailed and structured solutions used for SFT. The process is structured, with the reasoning steps enclosed within <solutions> tags and the final answer within <answer> tags.

\begin{figure}[]
  \centering
  \includegraphics[width=\textwidth]{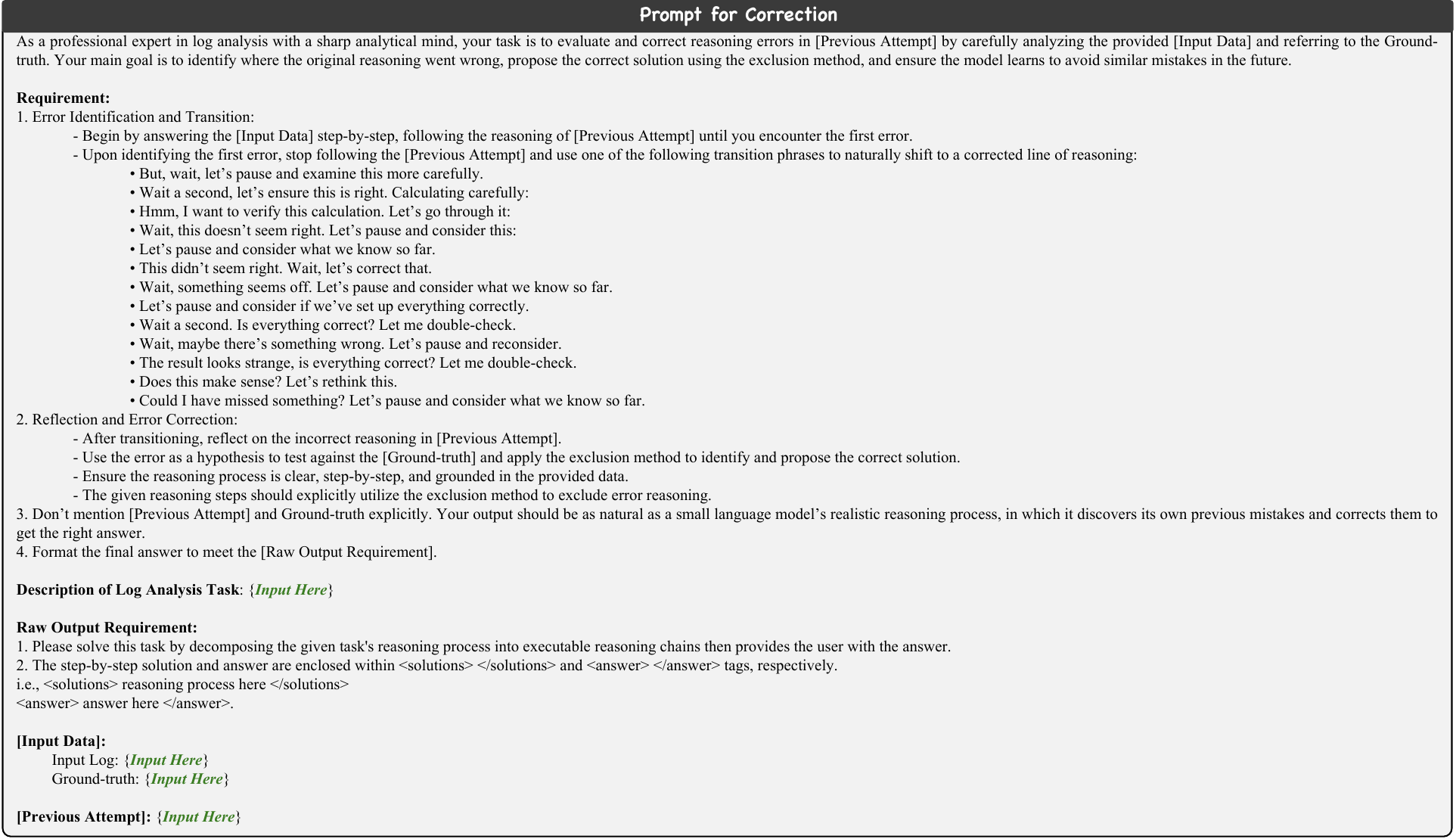}
  \caption{Prompt template for error correction.}
  \label{fig:logreason_correction_prompt}
\end{figure}

Figure \ref{fig:logreason_correction_prompt} shows the prompt template for correction, where this prompt is designed to guide a teacher LLM in identifying and correcting errors in a reasoning process. It focuses on evaluating a flawed reasoning trajectory, detecting the first error, and transitioning to a corrected solution using a systematic, exclusion-based approach. By reflecting on the mistakes, the expert reconstructs the reasoning chain step-by-step to ensure accuracy and improve the model's learning process. Then the corrected reasoning is used to build the preference data.

\end{document}